\documentclass[sigconf, nonacm]{aamas} 

\usepackage{balance} % for balancing columns on the final page
\usepackage{algpseudocode}
\usepackage{algorithm}  %,algorithmic}
\usepackage{xcolor}

\usepackage{dsfont}
\usepackage{amsthm}
\usepackage{amsmath}
\newtheorem{thm}{Theorem}
\newtheorem{mydef}{Definition}

\newtheorem{cor}{Corollary}

\newcommand{\paren}[1]{\mathopen{}\left( {#1}_{{}_{}}\,\negthickspace\right)\mathclose{}}
\newcommand{\bracket}[1]{\mathopen{}\left[ {#1}_{{}_{}}\,\negthickspace\right]\mathclose{}}
\newcommand{\brac}[1]{\mathopen{}\left\{ {#1}_{{}_{}}\,\negthickspace\right\}\mathclose{}}

\everymath{\displaystyle}
\def\S{\mathcal{S}}
\def\A{\mathcal{A}}

\def\G{\mathcal{G}}
\def\E{\text{E}}

\def\vili{\text{CVaRVILI}}
\def\viq{\text{CVaRVIQ}}
\def\forpec{\text{ForPECVaR}}

\title{Forward-PECVaR Algorithm: Exact Evaluation for CVaR SSPs}

\author{Willy Arthur Silva Reis, Denis Benevolo Pais, Valdinei Freire, Karina Valdivia Delgado}
\affiliation{
  \institution{University of São Paulo}
  \city{Sao Paulo}
  \country{Brazil}}
\email{willy.reis@usp.br, denis.pais@alumni.usp.br, valdinei.freire@usp.br, kvd@usp.br}

\begin{abstract}
The Stochastic Shortest Path (SSP) problem models probabilistic sequential-decision problems where an agent must pursue a goal while minimizing a cost function.
Because of the probabilistic dynamics, it is desired to have a cost function that considers risk. Conditional Value at Risk (CVaR) is a criterion that allows modeling an arbitrary level of risk by considering the expectation of a fraction $\alpha$ of worse trajectories.
Although an optimal policy is non-Markovian, solutions of CVaR-SSP can be found approximately with Value Iteration based algorithms such as CVaR Value Iteration with Linear Interpolation (\vili{}) and CVaR Value Iteration via Quantile Representation (\viq{}).
These type of solutions depends on the algorithm's parameters such as the number of atoms and $\alpha_0$ (the minimum $\alpha$). 
To compare the policies returned by these algorithms, we need a way to exactly evaluate stationary policies of CVaR-SSPs. Although there is an algorithm that evaluates these policies, this only works on problems with uniform costs.
In this paper, we propose a new algorithm, Forward-PECVaR (ForPECVaR), that evaluates exactly stationary policies of CVaR-SSPs with non-uniform costs.
We evaluate empirically CVaR Value Iteration algorithms that found solutions approximately regarding their quality compared with the exact solution, and the influence of the algorithm parameters in the quality and scalability of the solutions.
Experiments in two domains show that it is important to use an $\alpha_0$ smaller than the $\alpha$ target and an adequate number of atoms to obtain a good approximation. 
\end{abstract}


\newcommand{\BibTeX}{\rm B\kern-.05em{\sc i\kern-.025em b}\kern-.08em\TeX}

\begin{document}

%%% The following commands remove the headers in your paper. For final 
%%% papers, these will be inserted during the pagination process.

\pagestyle{fancy}
\fancyhead{}

%%% The next command prints the information defined in the preamble.

\maketitle 

%%%%%%%%%%%%%%%%%%%%%%%%%%%%%%%%%%%%%%%%%%%%%%%%%%%%%%%%%%%%%%%%%%%%%%%%

\section{Introduction}

\label{sec:intro}

The Stochastic Shortest Path (SSP) problem models probabilistic sequential-decision problems where an agent must pursue a goal while minimizing a cost function.
Because of the probabilistic dynamics, it is desired to have a cost function that considers risk. 
Several measures for measuring financial risk have been constantly studied and applied in different sectors. 
These measures include variance, \emph{Value at Risk} (VaR), and \emph{Conditional Value at Risk} (CVaR). 
Conditional Value at Risk (CVaR) is a coherent risk measure \cite{rockafeller:cvar-optimization} criterion that allows modeling an arbitrary level of risk by considering the expectation of a fraction $\alpha$ of worse trajectories in sequential-decision problems \cite{bauerle2011markov,chow:risk-sensitive,stanko2019risk}.

Although an optimal policy for CVaR-SSPs is non-Markovian (depends on the entire history of actions and states visited so far), solutions of CVaR-SSP can be found approximately with Value Iteration based algorithms.
One algorithm that finds a policy for CVaR-SSPs is the CVaR Value Iteration with Linear Interpolation, referred to as \emph{CVaRVILI} \cite{chow:risk-sensitive}. The policy returned by the CVaRVILI algorithm is stationary (it does not depend on time) and non-Markovian policy. This algorithm is very costly as it needs to solve several linear programming problems. Another algorithm that finds a stationary and non-Markovian policy is CVaR Value Iteration via Quantile Representation, referred to as \emph{CVaRVIQ} \cite{stanko2019risk}. CVaRVIQ uses distributional approach techniques to find the solutions faster than \vili{}.
The solutions that can be found approximately depend on the algorithm's parameters such as the number of atoms and $\alpha_0$ (the minimum $\alpha$).
Due to the approximate approach, the exact value of the policy is not found, although it is possible to approximate it as much as desired by using a greater number of atoms, which implies an increase in the computational cost.

To compare the policies returned by these algorithms, we need a way to exactly evaluate the stationary policies of CVaR-SSPs.  
Although there is an algorithm that evaluates these policies, this only works on problems with uniform cost \cite{Meggendorfer22}.
In this paper, we propose a new algorithm, 
ForPECVaR (Forward Policy Evaluation CVaR), that evaluates exactly stationary policies of CVaR-SSPs with non-uniform cost. To work with this type of cost, ForPECVaR tracks the accumulated cost for each trajectory from the initial state independently.

Thus, ForPECVaR could be used to compare approximate solutions. To the best of our knowledge, there is no algorithm that exactly evaluates the policies returned by \vili{}, \viq{}, and other similar algorithms for problems with non-uniform costs. 
ForPECVaR can also be used as the policy evaluation step of a possible Policy Iteration algorithm.
Additionally, \forpec{} also calculates the exact VaR value of the policies
that can be used by other algorithms such as the algorithm proposed in \cite{rigter2022planning}.

In this paper, we perform experiments to compare the approximate values returned by \vili{} and \viq{} with the exact values computed by ForPECVaR, and the influence of the algorithm parameters ($\alpha_0$ and number of atoms) in the quality and scalability of the solutions in two domains.

\section{Background}
\label{sec:background}

\subsection{Stochastic Shortest Path Problem}
A Stochastic Shortest Path Problem \cite{BertsekasTsitsiklis:1991} is described by a tuple $\mathcal{M}=\langle \mathcal{S},\mathcal{A},P,c,\mathcal{G} \rangle$ where: 
 $\S$ is a finite set of states;
 $\A$ is a finite set of actions;
$P:\S \times \A \times \S \rightarrow [0,1]$ is a transition function that represents the probability that $s'\in \S$ is reached after the agent executes an action $a \in \A$ in a state $s\in \S$, i.e., $\Pr(s_{t+1}=s'|s_{t}=s,a_{t}=a) = P(s,a,s')$;
 $c : \S \times \A \rightarrow \mathcal{R}^{+}$ is a positive cost function that represents the cost of executing an action $a \in \A$ in a state $s \in \S$, i.e., $c_{t}=c(s_{t},a_{t})$; and
 $\G$ is a non-empty set of goal states that are absorbing, i.e.,  $P(s_{t+1}\in\mathcal{G}|s_{t}\in\mathcal{G},a_{t}=a)=1$ and $ c(s_{t} \in\mathcal{G},a_{t}=a) =0$ for all $a\in\mathcal{A}$.
 
The solution to an SSP is a policy $\pi$ that could be:
\begin{itemize}
    \item Stationary ($\pi:\S\to\A$) that maps every state $s_{t}$ into an action $a_{t}=\pi(s_{t})$;
    \item Non-Markovian or history-dependent ($\Pi_H$). Let $H_t=H_{t-1} \times \A \times \S$ be the space of histories up to time $t \geq 1$ and $H_0=\S$ where each history $h_t \in H_t$ is $h_t= (s_0, a_0, \cdots, s_{t-1}, a_{t-1},s_t)$. 
    Let $\Pi_{H,t}$ be the set of all history-dependent policies up to time $t$ with the property that at each time $t$ the action is a function of $h_t$, i.e., $\Pi_{H,t}=(\mu_0: H_0 \rightarrow \A, \mu_1:H_1 \rightarrow \A, \cdots , \mu_t:H_t \rightarrow \A)$, then $\Pi_H = \lim_{t\rightarrow\infty} \Pi_{H,t}$ the set of all history-dependent policies.
\end{itemize}

Let the random variable $Z_{M}=\sum_{t=0}^{M}c\paren{s_{t},\pi(s_{t})}$ be the accumulated cost from time $0$ up to time $M$. The value function of a policy $\pi$ is defined by the total expected cost of reaching the goal from $s_0$:  
\begin{equation}
\begin{aligned}
V^{\pi}(s) = \lim_{M\to \infty} \E\left[ \left. Z_{M} \right| \pi,s_{0}=s \right].
\end{aligned}
\end{equation}

A policy $\pi$ is proper if $\lim_{t\rightarrow\infty}\Pr(s_{t}\in\G|\pi,s_{0}=s)=1$, $\forall{s\in\mathcal{S}}$ \cite{BertsekasTsitsiklis:1991}.
The value function $V^{\pi}(s)$ for SSPs is well-defined only for proper policies.
Any improper policy has an infinite value $V^\pi(s) = \infty$ at every state $s$ that cannot reach a goal state with probability 1.

The value of a proper policy $\pi$ can be found using the equation:
\begin{equation}
V^\pi(s) = \begin{cases}
0 & \text{, if }s\in \mathcal{G}\\
 c(s,\pi(s))+\displaystyle\sum_{s'\in\mathcal{S}}T(s,\pi(s),s')V^{\pi}(s') & \text{, otherwise.}
\end{cases}
\label{eq:VdeSSP}
\end{equation}

The optimal value $V^*(s)=\displaystyle\min_{\pi} V^\pi(s)$ can be computed by solving the Bellman equation:
\begin{equation}
V^*(s) = \begin{cases}
0 & \text{, if }s\in \mathcal{G}\\
 \displaystyle\min_{a\in \A} \left[c(s,a)+\displaystyle\sum_{s'\in\mathcal{S}}T(s,a, s')V^{*}(s')\right] & \text{, otherwise.}
\end{cases}
\label{eq:otimo:VdeSSP}
\end{equation}

An optimal policy can be extracted from $V^*$ by:
\begin{equation}
\pi^*(s) \in \displaystyle\arg\min_{a\in \A}\left[c(s,a)+ \displaystyle\sum_{s'\in \S}T(s,a,s') V^*(s') \right].
\label{eq:otimo:PideSSP}
\end{equation}

\subsection{VaR and CVaR metrics}

The VaR (\emph{Value at Risk}) and CVaR (\emph{Conditional-Value-at-Risk}) metrics are 
widely used for portfolio management of financial assets.
VaR measures the worst expected loss within a given $ \alpha $ confidence level, where $\alpha \in (0, 1)$.
VaR is defined as the $ 1- \alpha$ quantile of Z, i.e.:
%\begin{equation}
$VaR_\alpha(Z) = \min \{ z | F(z) \geq 1-\alpha \}$, 
%\end{equation}
where Z is a random variable (the accumulated cost in SSPs) and $F(z)$ is the cumulative distribution function. 

An alternative measure is the CVaR metric which is computed by averaging losses that exceed the VaR value. CVaR, with a confidence level $\alpha \in (0, 1)$ is defined as:
\begin{equation}
CVaR_\alpha(Z) = \min_{w\in\mathbb{R}} \brac{ w + \frac{1}{\alpha} \mathbb{E}\bracket{(Z-w)^{+}}},
\end{equation}
where  $(x)^{+}$ = max$\{x,0\}$ represents the positive part of $x$,
and $w$ represents the decision variable that, at the optimum point, reaches the value of VaR, i.e.,
%\begin{equation}
$CVaR_\alpha(Z) = VaR_\alpha(Z) + \frac{1}{\alpha} \mathbb{E}\bracket{(Z-VaR_\alpha(Z))^{+}}.$
%\end{equation}

There is a dual representation of CVaR that is used in sequential decision-making problems, defined as:
\begin{equation}
CVaR_{\alpha}(Z) =\max_{\xi  \in \mathcal{U}_{CVaR} (\alpha, \mathbb{P}) } \mathbb{E}_{\xi}[Z],
\end{equation}
where 
$\mathbb{E}_{\xi}[Z]$ is the $\xi$-weighted expectation of  Z,  $\mathcal{U}_{CVaR}$ is the risk envelope \cite{chow:risk-sensitive} that is represented by:
\begin{align}
 \mathcal{U}_{CVaR}(\alpha, \mathbb{P}) = \Bigg\{ \xi: \xi(\omega) \in &\bracket{0, \frac{1}{\alpha}}, \int_{\omega \in \Omega} \xi(\omega)\mathbb{P}(\omega)d\omega = 1 \Bigg\}, \label{eq:ucvar}
\end{align}
where $\mathbb{P}$ is a probability measure and $\Omega$ is the sample space. 
The risk envelope can be viewed as a set of probability measures that provides alternatives to $\mathbb {P}$ \cite{chow:risk-sensitive}.

\subsection{CVaR Stochastic Shortest Path Problem}

A CVaR SSP \cite{chow:risk-sensitive} is defined by the tuple $\mathcal{M}_{CVaR}= \langle \mathcal{M},\alpha\rangle$ where $\mathcal{M}$ is an SSP and $\alpha \in (0, 1]$ is the confidence level.
Remember that $\Pi_H$ is the set of history-dependent policies.
The objective in CVaR SSPs is to find $\mu\in{\Pi_H}$ \cite{chow:risk-sensitive}:
\begin{equation}
\label{prob:mincvar}
\min_{\mu \in \Pi_H}CVaR_{\alpha }\Bigg ( \sum_{t=0}^{\infty}  c(s_t,a_t) | s_0 = s , \mu \Bigg),
\end{equation}
where $\mu = \{\mu_0, \mu_1,...\}$ is the policy sequence that depends on the history with actions $a_t = \mu_t(h_t)$ for $t \in \{0,1,...\}$.

A dynamic programming formulation for the CVaR SSP problem was proposed by Chow et al. 
\cite{chow:risk-sensitive} by defining the CVaR value function $V$ over an augmented state space $\S \times Y$, where $Y=(0, 1]$ is a continuous confidence level.
\begin{equation}
V(s,y) = \min_{\mu \in \Pi_H}CVaR_{y}\Bigg ( \sum_{t=0}^{\infty}  c(s_t,a_t) | s_0 = s , \mu \Bigg).
\label{eq:value_augmented}
\end{equation}

The Bellman operator $T: \S \times Y \rightarrow \S \times Y$ is defined by \cite{chow:risk-sensitive}:
%\begin{equation}
\begin{align}
\nonumber
T&[V]\big(s,y\big) = \min_{a\in\A} \Bigg [c(s,a) + \\ 
%\nonumber
&\gamma \max_{\xi  \in \mathcal{U}_{CVAR} (y, P(.|s,a)) }	\sum_{s' \in \S} \xi(s')  V \Big (s', y\xi(s')\Big ) P(s'|s,a) \Bigg],
\label{eq:bellmanCvar}
\end{align}
%\end{equation}
where the risk envelope $\mathcal{U}_{CVAR}$ is defined in Eq. \ref{eq:ucvar}. This operator has two properties: contraction and concavity preserving in $y$ \cite{chow:risk-sensitive}.
The solution of $T[V](s,y)=V(s,y)$ is unique and equals to $V^*(s,y) = \min_{\mu \in \Pi_H}CVaR_{y}\Bigg ( \sum_{t=0}^{\infty}  c(s_t,a_t) | s_0 = s , \mu \Bigg)$.
The optimal policy can be obtained by a stationary Markovian policy, over the augmented state, defined as a greedy policy with respect to the value function $V^*(s,y)$.

The connection between an optimal history-dependent policy and a Markovian optimal policy on the augmented state space is given by the following theorem:

\begin{thm}
\textbf{(Optimal Policies \cite{chow:risk-sensitive})}
Let $\pi^*_H=\{\mu_0,\mu_1,\cdots\} \in \Pi_{H}$ a history-dependent policy recursively defined as:
\begin{equation}
    \mu_k(h_k)=u^*(s_k,y_k), \forall{k \geq 0},
\end{equation}
with initial conditions $s_0$ and $y_0=\alpha$, and augmented state transitions 
\begin{equation}
\nonumber
s_k \sim  P(\cdot|s_{k-1},u^*(s_{k-1},y_{k-1})), \quad y_k=y_{k-1}\xi^*_{s_{k-1},y_{k-1},u^*}(s_k), \forall{k \geq 1},
\end{equation}
where the stationary Markovian policy $u^*(s, y)$ and risk factor $\xi^*_{s,y,u^*}(\cdot)$ are solutions to the min-max optimization problem in the CVaR Bellman operator $T[V^*](s, y)$. Then, $\pi^*_{H}$ is an optimal policy for the CVaR SSP problem (\ref{prob:mincvar}) with initial state $s_0$ and CVaR confidence level $\alpha$.
\label{thm:optimal_policy}
\end{thm}

Among the algorithms that solve CVaR SSPs are \vili{} and \viq{}. 
\subsubsection{{\textbf \vili{} algorithm} \cite{chow:risk-sensitive}}
This algorithm makes a discretization of $Y$ creating a set of interpolation points (also called a set of interpolation atoms) and interpolates the value function among these points.
Let $N$ be the number of interpolation points (atoms) and for all $s \in \S$, let $Y(s) = (y_1, y_2, . . . , y_{N(s)} ) \in [0, 1]^{N(s)}$ be the set of interpolation points. The linear interpolation of  $y V(s,y)$ on these points is defined by:
%\begin{small}
\begin{equation}
\nonumber
I_s[V](y) = \begin{cases}
y_i V(s,y_i)  %&
\hspace{4cm} \text{, if }y \in Y\\
 y_i V(s,y_i)+ \frac{y_{i+1} V(s,y_{i+1}) - y_i V(s,y_i)}{y_{i+1} - y_i}(y - y_i), \\
\hspace{5.4cm} \text{otherwise,}
\end{cases}
\end{equation}
%\end{small}
where $y_i = \max\{y' \in Y(s) : y' \leq y\}$ and $y_{i+1}= \min\{y' \in Y(s) : y' \geq y\}$ such that $y\in$ $[y_i, y_{i+1}]$; i.e, we use the two nearest points of $y$, called $y_i$ and $y_{i+1}$. 
Since $I_{s'}[V] \big (y  \xi(s') \big )$ is the linear interpolation of $y \xi(s') V \big (s', y \xi(s') \big)$, in Eq. \ref{eq:bellmanCvar} we can replace $\xi(s')V\big (s', y  \xi(s')\big )$ by $\frac{I_{s'}[V] (y  \xi(s'))}{y}$, obtaining the following interpolated Bellman operator $T_I$ \cite{chow:risk-sensitive}:
\begin{align}
\nonumber
% T_{I}[Q]\big(s, y, a\big)
Q\big(s, y, a\big) = c(s,a) + \gamma \max_{\xi  \in \mathcal{U}_{CVAR} (y, P(.|s,a)) }	\sum_{s' \in S}  \frac{I_{s'}[V] \big (y \xi(s') \big )}{y} P(s'|s,a)
% \label{eq:bellmanCvarInterpolatedQ}
\end{align} 
\begin{align}
\centering
\nonumber
T_{I}[V]\big(s, y\big) = & \min_{a\in{A}} \{Q(s,y,a)\}.
\label{eq:bellmanCvarInterpolated}
\end{align}

CVaRVILI algorithm can have a high computational cost due to the need to solve many linear programming problems \textcolor{black}{($|S|\times|Y|\times|A|$ solver calls for each iteration)}. The greater the number of interpolation points $|Y|$, the smaller the approximation error, but the larger the computational time \cite{chow:risk-sensitive}.

\subsubsection{{\textbf\viq{} algorithm} \cite{stanko2019risk}} 
\textcolor{black}{This algorithm is} inspired by the use of the distributional approach of Bellemare et al \cite{bellemare2017distributional}.
The connection between the function $yCVaR_y$ and the quantile function ($VaR_y$) of the distribution of $Z$ are given by the convexity and piecewise linear properties of $yCVaR_y$ function, which means that $yCVaR_y$ can be obtained by: 
\begin{equation}
    yCVaR_y(Z) = \int_0^y VaR_\beta(Z)d\beta.
\label{eq:int}
\end{equation}
Additionally, $VaR_y$ can be obtained by:
\begin{equation}
    \frac{\partial}{\partial Z}yCVaR_y(Z) = VaR_y(Z).
\label{eq:der}
\end{equation}

\viq{} algorithm uses these properties to make faster computations.
For each augmented state $(s,y)$ and for each action $a$, the distribution of the values of each successor augmented state is extracted with the application of Eq. \ref{eq:der}
and then combined in another distribution considering the transition probability of each successor.
Finally, the new distribution is transformed in the value function $Q(s,y,a)$ with Eq. \ref{eq:int}.

The implementation of the CVaRVIQ algorithm returns the policy, CVaR, and VaR of all augmented states and does not return $\xi$. The $\xi$ function (necessary to guide the process of the variable $y$) can be computed by \cite{stanko2019risk}:
\begin{equation}
    \xi^\pi(s,y,s')=\frac{F_{Z^\pi(s')}(VaR_y(Z^\pi(s)))}{y},
\label{eq:xi_viq}
\end{equation}
where $Z^\pi(s)$ corresponds to the distribution of cumulative cost from state $s$ by following policy 
$\pi$ and $F_{Z^\pi(s')}$ to the cumulative distribution of $Z^\pi(s')$.
Intuitively, $y' = y\xi^\pi(s,y,s')$ corresponds to the portion of the tail of distribution $Z^\pi(s')$ to the $CVaR_{y}(s)$ under policy $\pi$. % Q(s,y,a)

\section{ForPECVaR Algorithm}
\label{sec:forpecvar}

In this section, we propose the ForPECVaR   algorithm, which evaluates a proper policy. Before presenting this algorithm, we introduce Theorem~\ref{th:piValue}, which shows how the CVaR value of a policy $\pi$ can be expressed in a forward approach, instead of a backup operator.

\subsection{$\pi$-Value}

The ForPECVaR algorithm is based on Theorem~\ref{th:piValue}. In Theorem~\ref{th:piValue}, $P^{X,\pi}(s)$ is the probability of reaching a goal state paying at most $X$ when following policy $\pi$.
Intuitively, Theorem~\ref{th:piValue} indicates that the CVaR value of a policy $\pi$ for $\alpha= 1-P^{X,\pi}(s)$ can be calculated by the difference between the mean value ($\mathbb{E}[Z|s_{0}=s,\pi]$) and the expected value of the best cases with cost at most $X$ divided by the probability of not reaching a goal state paying at most $X$.

\begin{mydef}
The probability of reaching a goal state starting at $s_{0}=s$ after following policy $\pi$ and paying  at  most
$X$ is defined by
\begin{align}
\nonumber
P^{X,\pi}(s)
= \Pr(Z\leq X | s_{0}=s,\pi) =
\lim_{T\to\infty}\Pr\paren{\sum_{t=0}^{T-1}c_{t}\leq X|s_{0}=s,\pi}.
\end{align}
\label{def:prob}
\end{mydef}
$X$ plays the role of $VaR_{\alpha=1-P^{X,\pi}(s)}$, as it will divide the Z distribution into $P^{X,\pi}(s)$ best cases and $1-P^{X,\pi}(s)$ worst cases.

\begin{thm}
Let the random variable $Z=\lim_{T\to\infty}\sum_{t=0}^{T-1}c_{t}$ be the accumulated cost and $\pi$ be a proper policy. Let  $\mathcal{X}^{\pi}(s) = \{X\in\mathds{R}|\Pr(Z=X|s_{0}=s,\pi)>0\}$ be the set of  accumulated cost with nonzero probability. For an SSP, $\mathcal{X}^{\pi}(s)$ is countable\footnote{Remember that the set of states is finite and the cost function is deterministic.}.
For all $X\in\mathcal{X}^{\pi}(s)$, we define:
$$y(X)=1-P^{X,\pi}(s).$$

The CVaR value of a policy  $\pi$ of the augmented state $\paren{s,y(X)}$ can be computed by:

\begin{equation}
%\nonumber
V^{\pi}\big(s,
%&
y(X)\big) =
%\\
%&
\frac{\mathbb{E}[Z|s_{0}=s,\pi] - \mathbb{E}\left[Z|Z\leq X,s_{0}=s,\pi\right]P^{X,\pi}(s)}{1-P^{X,\pi}(s) }. 
\label{resultTheorem}
\end{equation}
\label{th:piValue}
\end{thm}
\begin{proof}
Note that, because of the definition of $y(X)$, we have:
%\begin{equation}
\begin{align}
%\nonumber
V^{\pi}(s,y(X)) & = CVaR_{\paren{\alpha=y(X)}}(Z|s_{0}=s,\pi) 
%\\& 
= \mathbb{E}[Z|Z > X,s_{0}=s,\pi],
\label{eq:cvar}
\end{align}
%\end{equation}
and using the Law of Total Probability, we have:
\begin{align}
\nonumber
\mathbb{E}[Z|s_{0}&=s,\pi]  = %\\
%\nonumber
%&
\mathbb{E}[Z|Z> X,s_{0}=s,\pi]\Pr(Z > X|s_{0}=s,\pi) \\
\nonumber
&+ \mathbb{E}[Z|Z \leq X,s_{0}=s,\pi]\Pr(Z \leq X|s_{0}=s,\pi)
\end{align}
and implies:
\begin{align}
\nonumber &\mathbb{E}[Z|Z> X,s_{0}=s,\pi] = \\
&\frac{\mathbb{E}[Z|s_{0}=s,\pi] - \mathbb{E}[Z|Z \leq X,s_{0}=s,\pi]\Pr(Z \leq X|s_{0}=s,\pi)}{ \Pr(Z > X|s_{0}=s,\pi) }. \label{eq:manipulation}
\end{align}

Using Eqs.~\ref{eq:cvar} and \ref{eq:manipulation}, and the Definition~\ref{def:prob}, we obtain Eq. \ref{resultTheorem}. 
\end{proof}

\begin{cor}
Let $X\in\mathcal{X}^{\pi}(s)$ such that $y(X)$ is the greater value lesser or equal than $\alpha$  and $V^{\pi}\big(s,y(X)\big)$ the  CVaR value of a policy  $\pi$ of the augmented state $\paren{s,y(X)}$. The CVaR value of a policy  $\pi$ of the augmented state $\paren{s,\alpha}$ can be computed by:

\begin{align}
V^{\pi}(s,\alpha) \gets
\displaystyle\frac
{y(X) V^{\pi}(s,y(X)) + (\alpha-y(X)) X}
{\alpha}.
\label{eq:cor}
\end{align}

\label{cor:initialAugmentedStateValue}
\end{cor}

\subsection{ForPECVaR Algorithm}

The ForPECVaR algorithm (Algorithm~\ref{alg:forpecvar}) makes use of Theorem~\ref{th:piValue} to compute CVaR values $V^\pi(s_0,\alpha)$ for a proper policy $\pi$ and an initial state $s_0$ considering a target $\alpha$. 
The ForPECVaR algorithm constructs a tree from the initial augmented state ($s_{0},\alpha$) and expands leaves until a goal state is reached. Leaves with the smallest accumulated cost are expanded first, so that the minimum cost trajectory is founded first. 
Globally, the ForPECVaR algorithm keeps the expected value of the best cases with cost at most $X$, i.e., 
$\mathbb{E}\left[Z|Z\leq X,s_{0}=s,\pi\right]$  (represented in the algorithm by $V_{\leq X}$), and $P^{X,\pi}(s_0,s')$
 (represented in the algorithm by $P^{X,\pi}$).

The algorithm has as input an SSP MDP $\mathcal{M}$, a proper policy $\pi$, an initial state $s_0$, a target $\alpha$ and an admissible heuristic function $heuristic$\footnote{
A function $heuristic$ is admissible if $heuristic(s) \leq V^*(s), \forall s$.}.
The initial state $s_0$ and the target $\alpha$ can be interpreted as the initial augmented state $(s_0,\alpha)$ that originates the trajectories. 
The algorithm assumes that there is an algorithm capable of evaluating the policy $\pi$ with risk-neutral criteria.

The priority queue used in the algorithm is formed by nodes that include the following information: the state ($s$), its accumulated cost ($cost$), the probability of the node being reached ($prob$), execution history ($h$), current stage ($t$), and priority ($priority$).
The queue is initialized with node $d$ corresponding to the initial state ($s_0$) with accumulated cost $d.cost=0$, current stage $d.t=0$, probability of being reached $d.prob=1$, history  $d.h \gets \{s_0\}$ and priority $d.priority=0$ (lines 3 and 4).

In line 5, $\mathbb{E}[Z|s_{0}=s,\pi]$ (the risk-neutral value function of $\pi$, represented in the algorithm by $V^{\pi}_{ mean}$) is computed. 
From $s_0$, the tree is expanded until the probability of reaching a goal state starting at $s_{0}$ after following policy $\pi$ and paying at most $X$ is greater than or equal to $1-\alpha$ (lines 6 to 26).
At each iteration, the node with the highest priority is removed from the queue (line 7). Similar to the A* algorithm, we consider the lowest accumulated cost of the state plus the admissible heuristic as the highest priority.
If the removed node $d$ is not a goal state, the action is taken from the policy, and the nodes corresponding to the successor augmented states of $d$ are queued (lines 8 to 18). In line 14, the action, the cost of applying that action in the state $d.s$ and the next state $s'$ are concatenated in the history.  
In line 17, a new node is inserted in the queue if there is no other node in it with the same policy augmented state\footnote{Policy augmented state represents a state $s$ in the case of a Markovian policy; an augmented state $(s,\alpha)$ in the case of a CVaR policy; and a history in the case of a non-Markovian policy, which means that the node is unique.} and accumulated cost. Otherwise, the data is grouped with an existing equivalent node. In this case, the probability of the state is added to the existing probability already found.

Once a goal state is reached, the variables that depend on $X$ can be updated: 
(1) the expected value of the best cases with cost at most $X$ (line 20); 
(2) the probability of reaching a goal state starting at $s_{0}$ after following policy $\pi$ and paying at most $X$ (line 21); 
(3) the corresponding $\alpha$, represented by $y^X$ (line 22); and 
(4) the value function of the augmented state $V^\pi(s_0,y^X)$ which is calculated according to Eq.~\ref{resultTheorem} (line 23).

In line 27, the algorithm uses the values of the last $y^X$ and accumulated cost $X$ to calculate the value of $V^\pi(s_0,\alpha)$ according to Eq.~\ref{eq:cor}, which is returned by the algorithm in line 28, along with X. Note that X value is the $VaR_\alpha(s_0)$, which can be used in other algorithms, as will be discussed in the Related Work section.

\begin{algorithm}[!t]
\caption{{\sc ForPECVaR}}
\footnotesize
\begin{algorithmic}[1]
 
% linha 1 - Input
\State \textbf{Input:} an SSP $\mathcal{M}=\langle \mathcal{S},\mathcal{A},P,c,\gamma,\mathcal{G} \rangle$, a policy $\pi$, a state $s_{0}$, a target $\alpha$ and a heuristic function $heuristic$ 

\State $P^{X,\pi} \leftarrow 0, V_{\leq X} \leftarrow 0$

\State $q\gets \emptyset$ \Comment{PriorityQueue}

\State $q.insert(d.s \gets s_0, d.cost \gets 0, d.prob \gets 1, d.h \gets (s_0), d.t \gets 0, d.priority \gets 0)$ 
\label{forpec:insert_s0}

\State $V^{\pi}_{mean} \gets {\sc MDPPolicyEvaluation}(\mathcal{M},\pi)$
\label{forpec:policy_evaluation}

\State \textbf{while} $1-P^{X,\pi} > \alpha$ %

\State \hspace{\algorithmicindent} $d\gets q.getHighestPriorityState()$ 

\State \hspace{\algorithmicindent} \textbf{if} $d.s\notin\mathcal{G}$

 \State \hspace{\algorithmicindent}\hspace{\algorithmicindent}$a \gets \pi(d.h)$
 \label{forpec:actionExtraction}

 \State \hspace{\algorithmicindent}\hspace{\algorithmicindent}\textbf{for} each next state $s'$ from $d.s$ following a \textbf{do}
 \label{forpec:for_begin}
 
 \State \hspace{\algorithmicindent}\hspace{\algorithmicindent}\hspace{\algorithmicindent} $data.s\gets s'$
 
 \State \hspace{\algorithmicindent}\hspace{\algorithmicindent}\hspace{\algorithmicindent} $data.cost\gets d.cost + \gamma^{d.t}c(d.s,a)$
 
 \State \hspace{\algorithmicindent}\hspace{\algorithmicindent}\hspace{\algorithmicindent} $data.prob \gets d.prob \times P(d.s,a,s')$
 
 \State \hspace{\algorithmicindent}\hspace{\algorithmicindent}\hspace{\algorithmicindent} $data.h\gets (data.h, a, c(d.s,a), s')$
 
 \State \hspace{\algorithmicindent}\hspace{\algorithmicindent}\hspace{\algorithmicindent} $data.t\gets d.t+1$
 
 \State \hspace{\algorithmicindent}\hspace{\algorithmicindent}\hspace{\algorithmicindent} $data.priority\gets d.cost + \gamma^{d.t} \times c(d.s,a) + \gamma^{{d.t}+1} \times heuristic(s')$
 
 \State \hspace{\algorithmicindent}\hspace{\algorithmicindent}\hspace{\algorithmicindent} $q.insertAndGroup(data)$
 
 \State \hspace{\algorithmicindent}\hspace{\algorithmicindent}\textbf{endfor}
 \label{forpec:for_end}

\State \hspace{\algorithmicindent}\textbf{else}

 \State\hspace{\algorithmicindent}\hspace{\algorithmicindent}$V_{\leq X} \gets \displaystyle\frac{V_{\leq X}\times P^{X,\pi}+d.cost \times d.prob}{P^{X,\pi}+d.prob}$
 
 \State\hspace{\algorithmicindent}\hspace{\algorithmicindent}$P^{X,\pi} \gets P^{X,\pi} + d.prob$
 
 \State\hspace{\algorithmicindent}\hspace{\algorithmicindent}$y^X \gets 1-P^{X,\pi}$
  
 \State\hspace{\algorithmicindent}\hspace{\algorithmicindent}$V^{X,\pi}(s_0,y^X) \gets  \displaystyle\frac{V_{mean}(s_0)-V_{\leq X}\times P^{X,\pi}}{y^X}$
  
  \State \hspace{\algorithmicindent}\hspace{\algorithmicindent}$X\gets d.cost$ 
 
 \State\hspace{\algorithmicindent} \textbf{endif} 
 
\State \textbf{endwhile}

\State $V^{\pi}(s_0,\alpha) \gets
\displaystyle\frac
{y^X\times V(s_0,y^X) + (\alpha-y^X)\times X}
{\alpha}$

\State \textbf{return} $CVaR_\alpha(s_0)=V^{\pi}(s_0,\alpha)$
and $VaR_\alpha(s_0) = X$
\end{algorithmic}
\label{alg:forpecvar}
 
\end{algorithm}

\subsection{\forpec{} applied to \vili{} and \viq{} solutions}

Since the \forpec{} algorithm evaluates any proper policy that depends on history, 
this algorithm can be used to evaluate the solutions of \vili{} and \viq{} algorithms that depend on the augmented state space. Remember that the solution of the \vili{} algorithm is composed of the policy $\pi$ and the function $\xi$; and the solution of the \viq{} algorithm is composed of the policy $\pi$ and the function $VaR$, which can be used to implicitly obtain the function $\xi$ (Eq.~\ref{eq:xi_viq}). 
The function $\xi$ is used to calculate the process $y_{0}=\alpha,y_{1},y_{2},\dots$, governed by: 
\begin{equation}
  y_{t} = y_{t-1} \xi(s_{t-1},y_{t-1},s_{t}), 
  \label{eq:procxi}
\end{equation}
which is used to obtain the augmented state $(s_t, y_t)$ in stage $t$ of each trajectory.
In both algorithms, \vili{} and \viq{}, the solutions are discretized by the set of atoms $Y(s) = \{y_1, y_2, \ldots, y_{N(s)} \} \in [0, 1]^{N(s)}, \forall s\in{S}$.

Summing  up, to evaluate \vili{} and \viq{} solutions using \forpec{} (Algorithm \ref{alg:forpecvar}), we need to (i) define how to work with policies that depend on the augmented state space and are discretized, and (ii) how to perform the policy evaluation in line 5 of Algorithm \ref{alg:forpecvar}.

 \paragraph{{\bf Working with policies that depend on the augmented state space and are discretized}} 
Since the policy $\pi$ depends on the augmented state space, we need to store $\alpha$ instead of the history $h$ in the priority queue. Thus, line 9 of Algorithm \ref{alg:forpecvar}  must be replaced by:
$a \gets \pi(d.s, d.\alpha).$
Additionally,   
$d.h \gets \{s_0\}$ in line 4 must be replaced by:
$d.\alpha=\alpha$
and $data.h\gets (data.h, a, c(d.s,a), s')$ in line 14 must be replaced by:
\begin{align}
 \nonumber
    &\alpha_{Next} \gets d.\alpha \times \xi(d.s,d.\alpha,s'),\\
\nonumber    
    &\alpha' \gets \arg\min_{y\in{Y}} \{{abs(log(y)-log(\alpha_{Next}))},\\
\nonumber
    &data.\alpha \gets \alpha'.
\end{align}

The first assignment corresponds to the application of Eq.
\ref{eq:procxi}. In the second assignment, the discretization of the alpha value is done by obtaining the closest atom $y\in{Y}$, considering  the log distance.

Finally, the method insertAndGroup (line 17) inserts a new node in the queue if there is no other node in it with the same state, $\alpha$ and accumulated cost. Otherwise, the data is grouped with an existing equivalent node and the probability is also added to the previous probability.

\paragraph{{\bf Computing $V^{\pi}_{ mean}$}}

The policy evaluation in line \ref{forpec:policy_evaluation} of  \forpec{}  (Algorithm \ref{alg:forpecvar}) to  evaluate \vili{} and \viq{} solutions is defined in Algorithm \ref{alg:pe_ext}. This algorithm uses the classical policy evaluation algorithm \cite{puterman:book}. In our implementation, this method also returns an admissible heuristic function for the augmented states.

MDPPolicyEvaluation (Algorithm \ref{alg:pe_ext}) takes an SSP MDP, a proper policy $\pi$ and a function $\xi$ and calculates the value functions $V^\pi_{mean}$ and $V^ \pi_{min}$.
In line 2, the algorithm CreateExtendedMDP (Algorithm \ref{alg:create_extmdp}) is called to create the extended MDP $M^\pi$, which has a single action per augmented state.
The policy evaluation of the unique policy of $M^\pi$ is executed in line 3 by a classical policy evaluation algorithm \cite{puterman:book}.
The worst possible value for the policy $V^ \pi_{min}$ is calculated in line 4 considering the determinization of $M^\pi$ followed by its policy evaluation. 
This value function can be used as an admissible heuristic for the ForPECVaR algorithm.

\begin{algorithm}[!t]
 \caption{MDPPolicyEvaluation}
 \footnotesize
 \begin{algorithmic}[1]
 
 \State \textbf{Input}: $\mathcal{M}$, $Y$, $\pi$ and $\xi$ 
  
 \State $M^\pi \gets CreateExtendedMDP(\mathcal{M}, Y, \pi, \xi)$
 
 \State $V_{mean}^\pi \gets PolicyEvaluation(M^\pi)$
 
 \State $V_{min}^\pi \gets PolicyEvaluationDeterminizedMDP(M^\pi)$
 
 \State return $V^\pi_{mean}, V^\pi_{min}$ 
  \end{algorithmic}
  \label{alg:pe_ext}
 \end{algorithm}

CreateExtendedMDP (Algorithm \ref{alg:create_extmdp}) creates a new MDP from the MDP $\mathcal{M}$ considering the policy $\pi$ and function $\xi$ to obtain the new transition probability function.
In this new MDP, the set of states is defined by $X = \S \times Y$ (line 2), where each state $x$ represents an augmented state $(s,y)$.
For each possible augmented state $(s,y)$, the cost function is defined as the same cost of the state $s$ (line 12) in $\mathcal{M}$.
The $\alpha$ value of each successor state of $s$ is computed by the application of Eq.~\ref{eq:procxi}
using the function $\xi$ (line 7). Then, this $\alpha$ is approximated to the nearest atom in the set $Y$ considering the logarithmic distance (line 8).
The transition probability function is defined in line 10. 

  \begin{algorithm}[!t]
 \caption{CreateExtendedMDP}
 \footnotesize
 \begin{algorithmic}[1]
 
 \State \textbf{Input: $\mathcal{M}=\langle \mathcal{S},\mathcal{A},P,c,\gamma,\mathcal{G} \rangle$, $Y$, $\pi$ and $\xi$}
 
 \State $X \gets \mathcal{S} \times Y$
    
 \State $\textbf{for}~(s, y) \in \mathcal{S} \times Y$
 
 \State \hspace{\algorithmicindent} $x \gets (s,y)$ 
 
 \State \hspace{\algorithmicindent} $a \gets \pi(s, y)$
 
 \State \hspace{\algorithmicindent}  \textbf{for} each next state $s'$ from $s$ following $a$ \textbf{do}
 
 \State \hspace{\algorithmicindent} \hspace{\algorithmicindent} $\alpha_{Next} \gets y \xi(s,y,s')$
 
 \State \hspace{\algorithmicindent} \hspace{\algorithmicindent} $\alpha' \gets \arg\min_{y\in{Y}} \{{abs(log(y)-log(\alpha_{Next}))}\}$
 
 \State \hspace{\algorithmicindent} \hspace{\algorithmicindent} $x' = (s', \alpha')$
 
 \State \hspace{\algorithmicindent} \hspace{\algorithmicindent} $T(x,a,x') \gets P(s,a,s')$
 
 \State \hspace{\algorithmicindent} \textbf{endfor}
 
 \State \hspace{\algorithmicindent} $C(x,a) \gets C(s,a)$
 
  \State \textbf{endfor}

 \State return $M = \langle X, A = \mathcal{A}, T, C, G = \mathcal{G}\times{Y}\rangle$ 
  \end{algorithmic}
  \label{alg:create_extmdp}
 \end{algorithm}

\section{Experiments}
\label{sec:experiments}

 In this section, we compare CVaRVILI \cite{chow:risk-sensitive} and CVaRVIQ \cite{stanko2019risk}
  in terms of execution time and quality of the solution.
Both algorithms return the value function CVaR ({\bf \emph{approximate value}}) and the policy for all augmented states $(s,y) \in \mathcal{S}\times{Y}$. The quality of the policy is evaluated exactly with the proposed algorithm, ForPECVaR. In this section, we call  this computed value by ForPECVaR as {\bf \emph{exact value}}. 
With the experiments, we want to answer the following questions:
    (1) What are the differences between the CVaRVILI and CVaRVIQ algorithms in terms of the approximate value, exact value, and execution time?; and
    (2) What is the influence of CVaRVIQ parameters (number of atoms $|Y|$ and $\alpha_0$) on the approximate and exact values? Are there some insights about how to choose these parameter values for a problem?
 
We used a desktop machine running with 6 processors at 2.90 GHz and 24 GB of memory DDR4. We executed the experiments in the Gridworld domain used in  \cite{chow:risk-sensitive} and \cite{stanko2019risk}, 
and the River domain \cite{Freire:2017:GUS:3091125.3091231}. 
In both domains, we set $\epsilon = 0.001$ as the residual error and $\gamma=1$.
The parameters values used in the experiments are  $|Y| \in \{7,13,25\}$, where $|Y|=N(s), \forall s \in \S$, and  $\alpha_0\in\{10^{-3}, 10^{-2}, 10^{-1}\}$.

\paragraph{\bf Gridworld domain}
An agent moves in a grid with $m$ rows by $n$ columns 
from an initial location $(n,m)$ to a goal location $(1,m)$ through movements in the cardinal directions (N, S, E or W). Transitions occur in the direction of movements with 0.95 of chance but can happen in each other direction with the residual probability equally distributed. Transitions to invalid grid locations maintain the robot in the same location. Each movement has cost 1, except 
in the goal location where actions have zero cost. 
The grid has obstacles that simulate the end of the agent run with a transition to the goal state with a cost of 100.
We performed experiments with three problems: $5\times5$ with 25 states, $8\times9$ with 72 states, and $14\times16$ with 224 states.

\paragraph{\bf River domain}
The agent moves in a grid with $m$ rows (height) by $n$ columns (width) from an initial location on one bank of the river to a goal location on the other bank of the river. The agent moves in the cardinal directions (N, S, E or W).
The sides of the grid represent the banks, the top represents a bridge, and the bottom a waterfall. The other locations represent the river itself.
Transitions in the banks and the bridge are deterministic. 
If the agent falls into the waterfall, it is transported deterministically to the start location (initial state) at the first line above the waterfall on the left bank.
Transitions in the river occur in the direction of the action with probability $0.8$ and the agent stays in the same position with probability $0.2$ (movement $m_1$ with probability $p_1$). These transitions also depend on movement $m_2$ with probability $p_2$, which is equal to $0.2$  of falling down one level of the river (decrease one row) and $0.8$ of maintaining in the same position. The resulting position and transition probability are determined by the two movements. Thus, the transition probability in the river is $p_1  p_2$. Each deterministic action has cost 1, and  probabilistic actions have a cost of 2 to move north, 1 to move east and west, and 0.5 to move south. The actions applied at the goal state have zero cost.
We performed experiments with three problems: 
$10\times3$, $16\times6$, and $30\times10$ with  30, 96, and 300 states, respectively.

\subsection{Approximate and exact values of CVaRVILI and CVaRVIQ}

We calculated 810 approximate values and 810 exact values for the algorithms \vili{} and \viq{}  considering the 2 domains, 3 problems per domain, 3 values of $|Y|$, 3 values of $\alpha_0$ and initial augmented states $(s_0,y_i), \forall y_i\in{Y}$.
The difference between the approximate values obtained by \vili{} and \viq{} in all 810 points was less than $10^{-6}$.
The difference between exact values of both algorithms was less than $0.001$ in $97.28\%$ of the points, less than $0.01$ in $99.26\%$, and less than $0.1$ in $100\%$.

\begin{figure}[!t]
\centering
  \includegraphics[width=.23\textwidth]{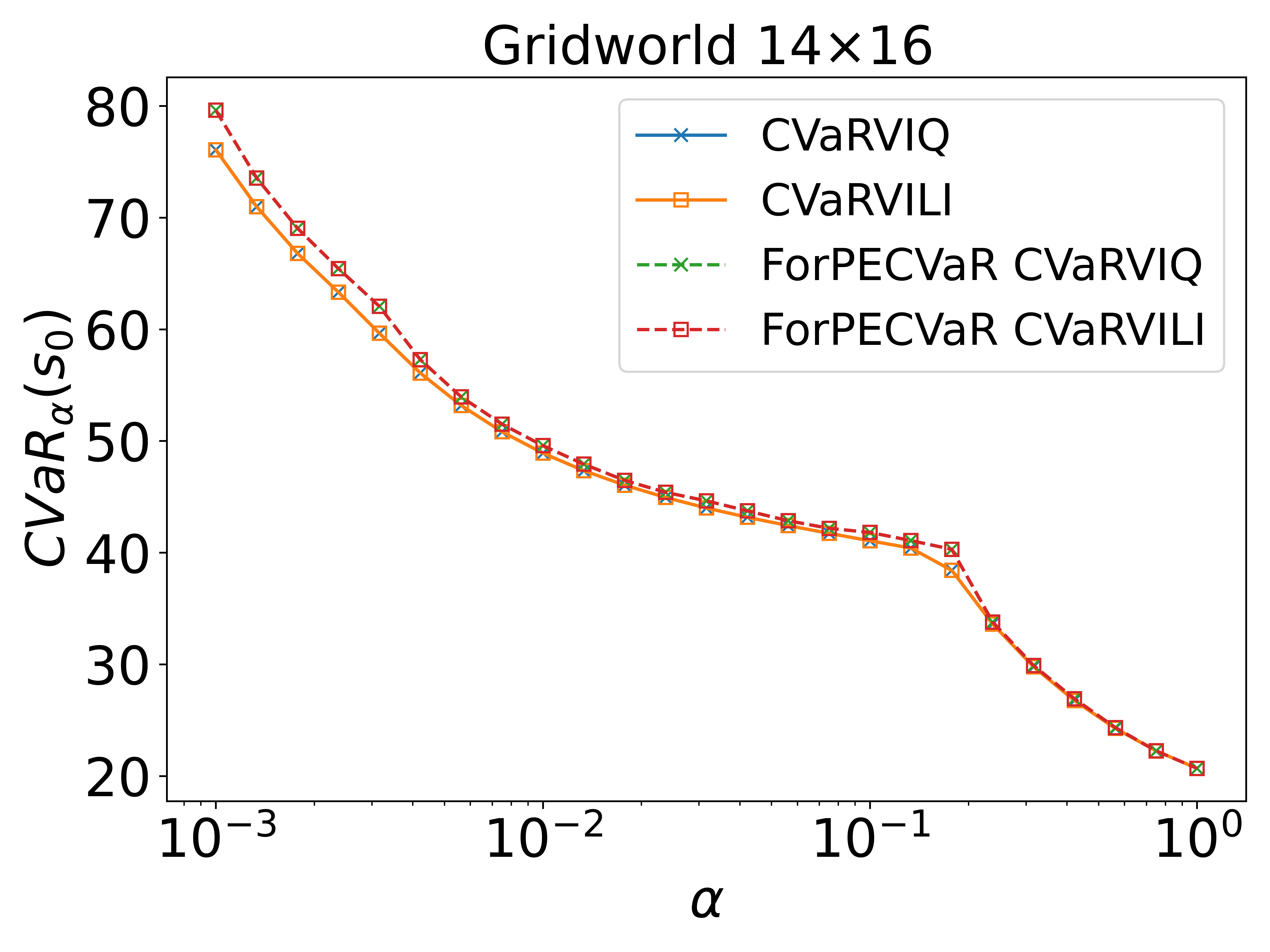}
  \includegraphics[width=.23\textwidth]{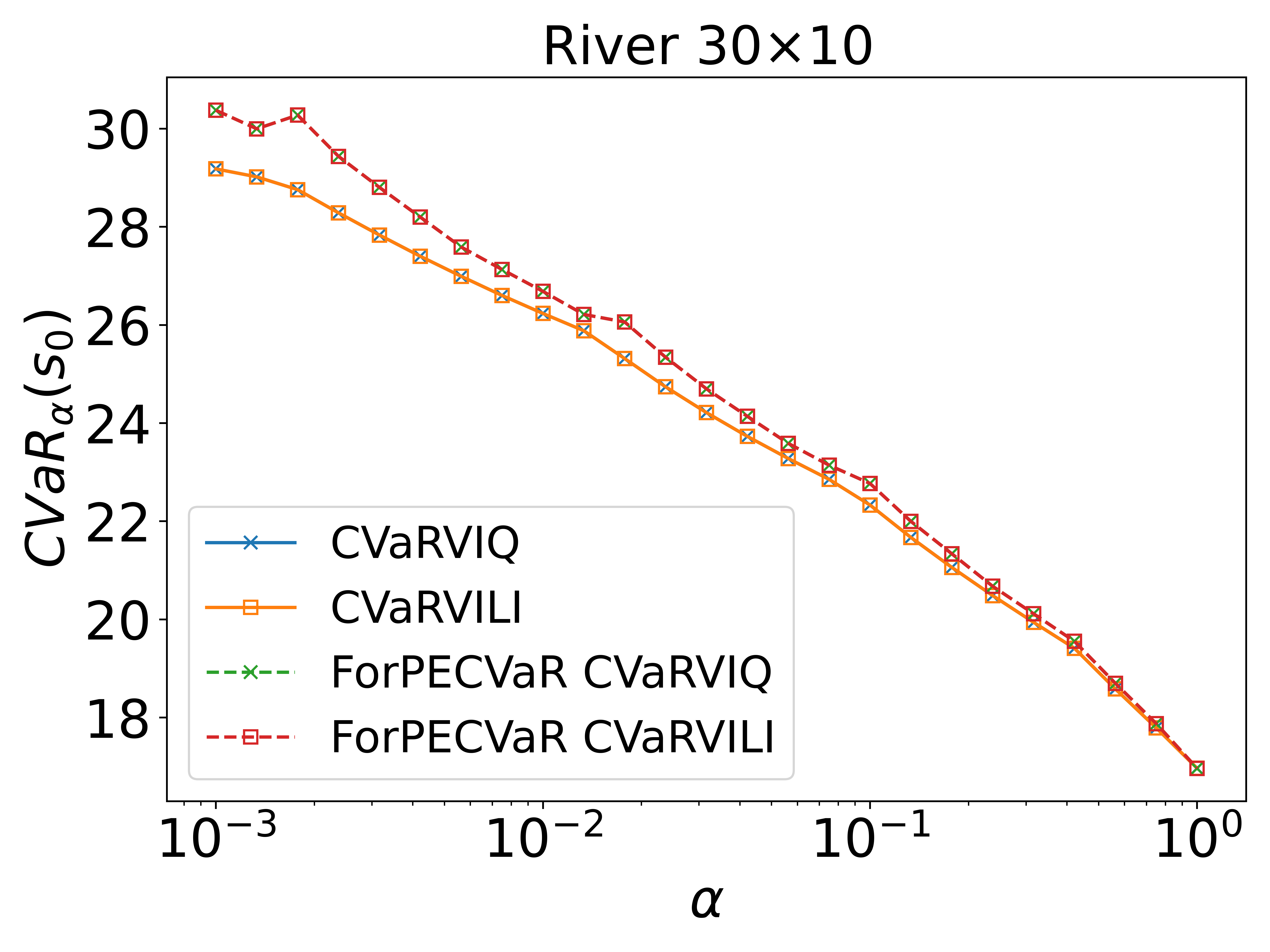}
\caption{\centering Approximate and exact values 
with $\alpha_0=10^{-3}$ and $|Y|=30$. 
}
\label{fig:forpecvar_both}
\end{figure}

Fig. \ref{fig:forpecvar_both} shows the approximate and the exact values for Gridworld $14 \times 16$ and River $30 \times 10$. 
Solid lines correspond to approximate values, while dashed lines correspond to exact values.
The corresponding approximate and exact value lines were paired at their markers.
Although the results of the solutions of \vili{} and \viq{} are close both in relation to the approximate and the exact values, the execution time of the algorithms is significantly different.
 \viq{}  is at least one order of magnitude faster than \vili{} (which needs to solve several linear programming problems) and can be up to two orders of magnitude faster in experiments with more atoms. 

\begin{figure}[!t]
\centering
  \includegraphics[width=.23\textwidth]{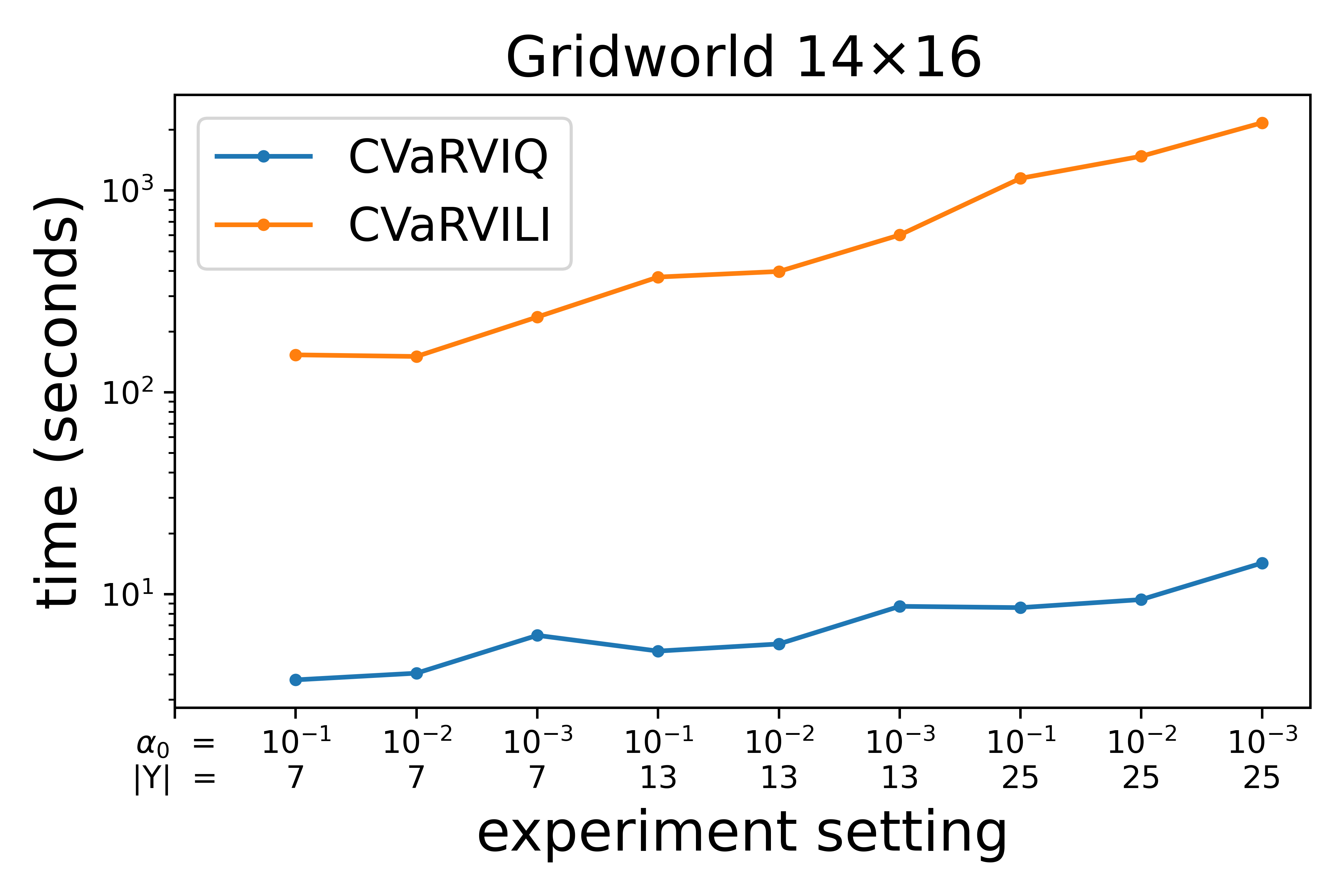}
  \includegraphics[width=.23\textwidth]{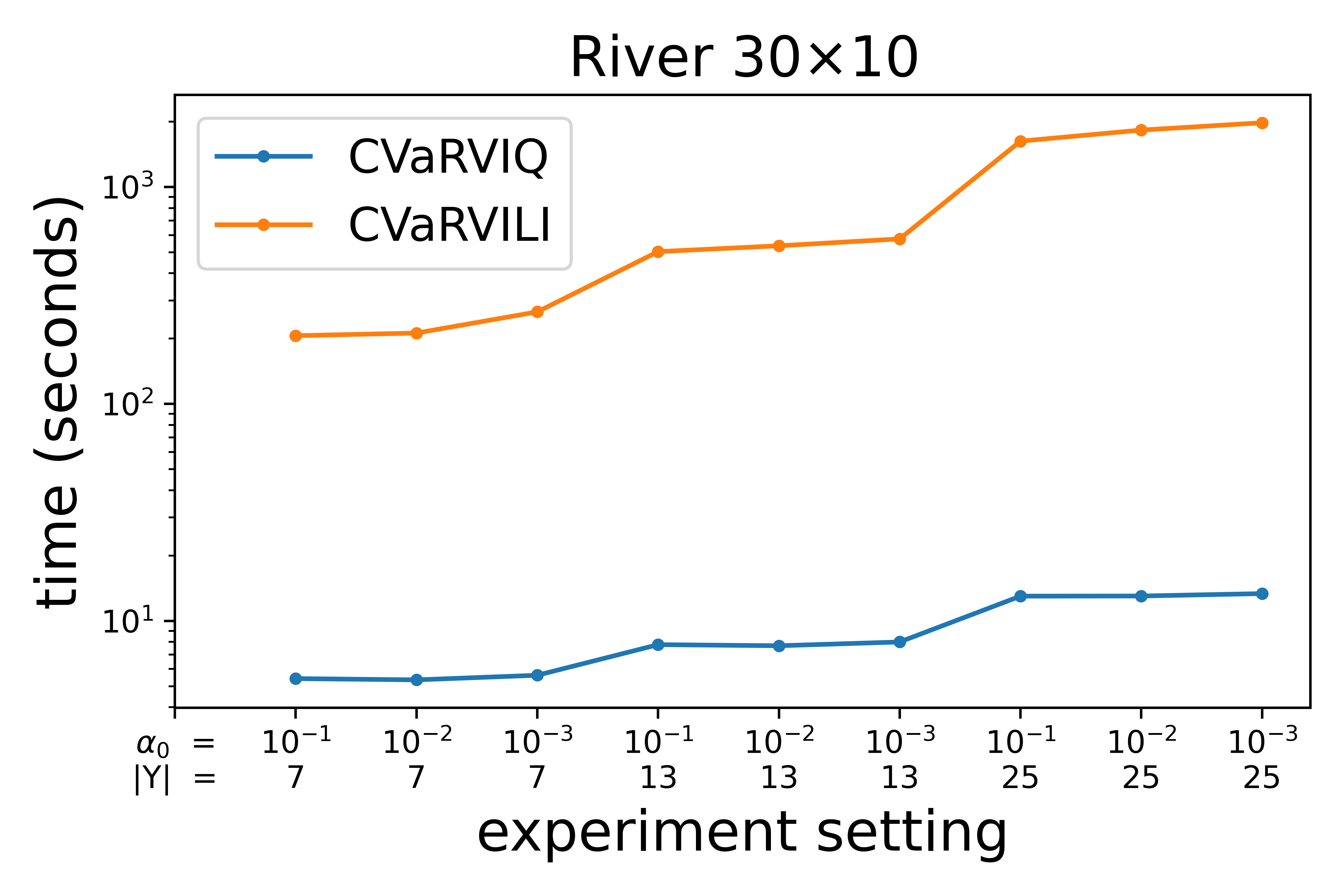}
\caption{\centering Execution time of \viq{} and \vili{}.}% algorithms.}
\label{fig:time_both}
\end{figure}

Fig. \ref{fig:time_both} shows the runtime in seconds for the Gridworld $14\times16$ and the River $30\times10$ problems for different settings varying $\alpha_{0}$ and $|Y|$.
The figure shows that with more atoms it takes longer to solve the problem.
The value of $\alpha_{0}$ influences the execution time, but to a lesser extent than the number of atoms.
The results of the other problems are similar.

Next, we analyze the influence of \viq{}'s  parameters on the quality of its solutions through two experiments.
In the first one, we fixed the parameter $\alpha_0$ and varied $|Y|$. 
In the second one, we fixed the parameter $|Y|$ and varied the $\alpha_0$ values.
The analysis is done only for the \viq{} algorithm, since both algorithms have similar results in terms of quality, and the \viq{} is more efficient than the \vili{} in terms of execution time.

\subsection{Values of CVaRVIQ varying $|Y|$}

Chow et al. \cite{chow:risk-sensitive}, in Theorem 7, show that the approximation error tends to be zero when the number of interpolation points is arbitrarily large by the application of the Interpolated Bellman operator. 
We believe that \viq{} has the same behavior since the approximate values of both are similar in the experiments performed. 
We analyzed this by varying the parameter $|Y|$ with $\alpha_0$ fixed.

\begin{figure}[!t]
\centering
  \includegraphics[width=.23\textwidth]{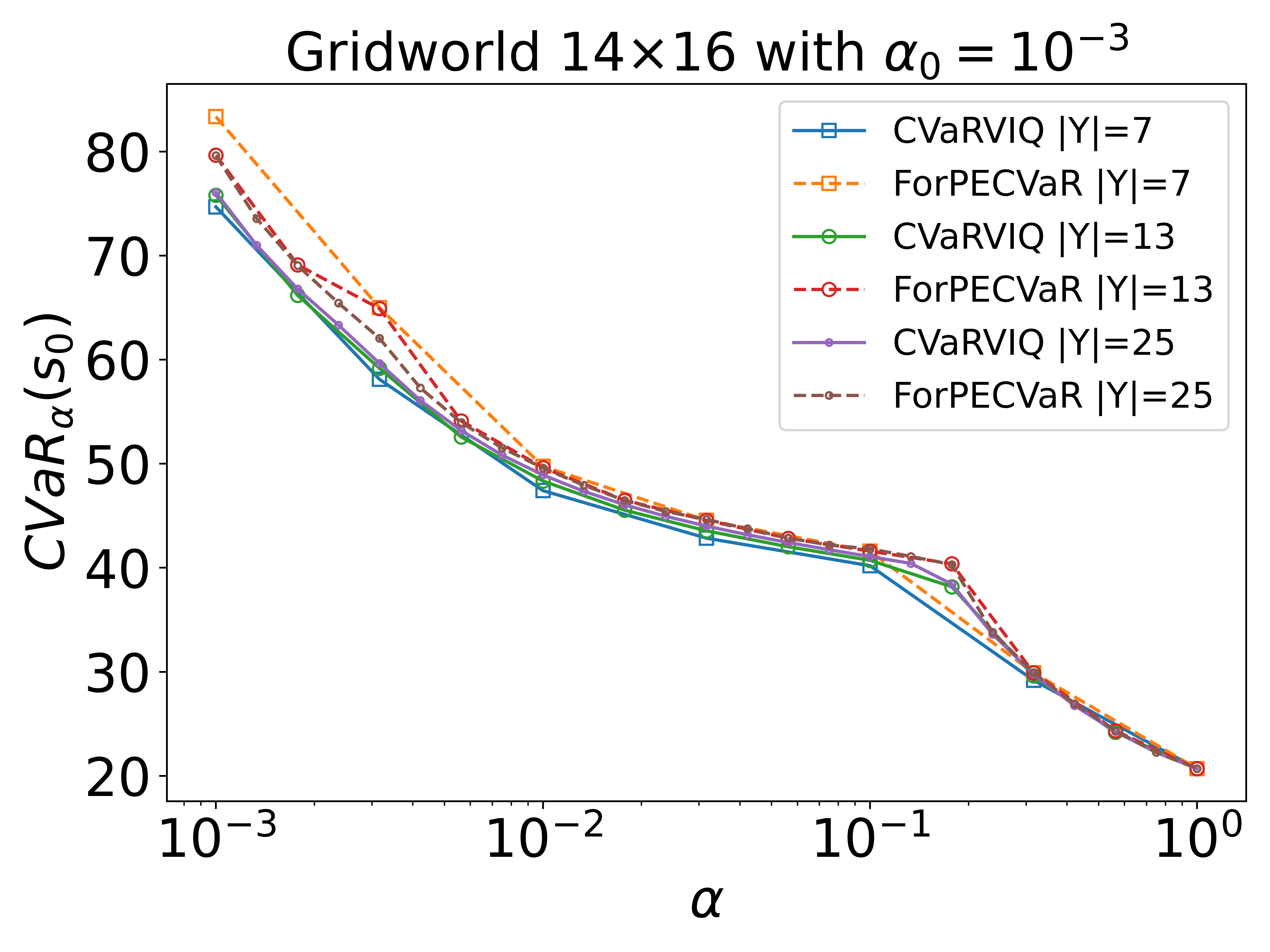}
  \includegraphics[width=.23\textwidth]{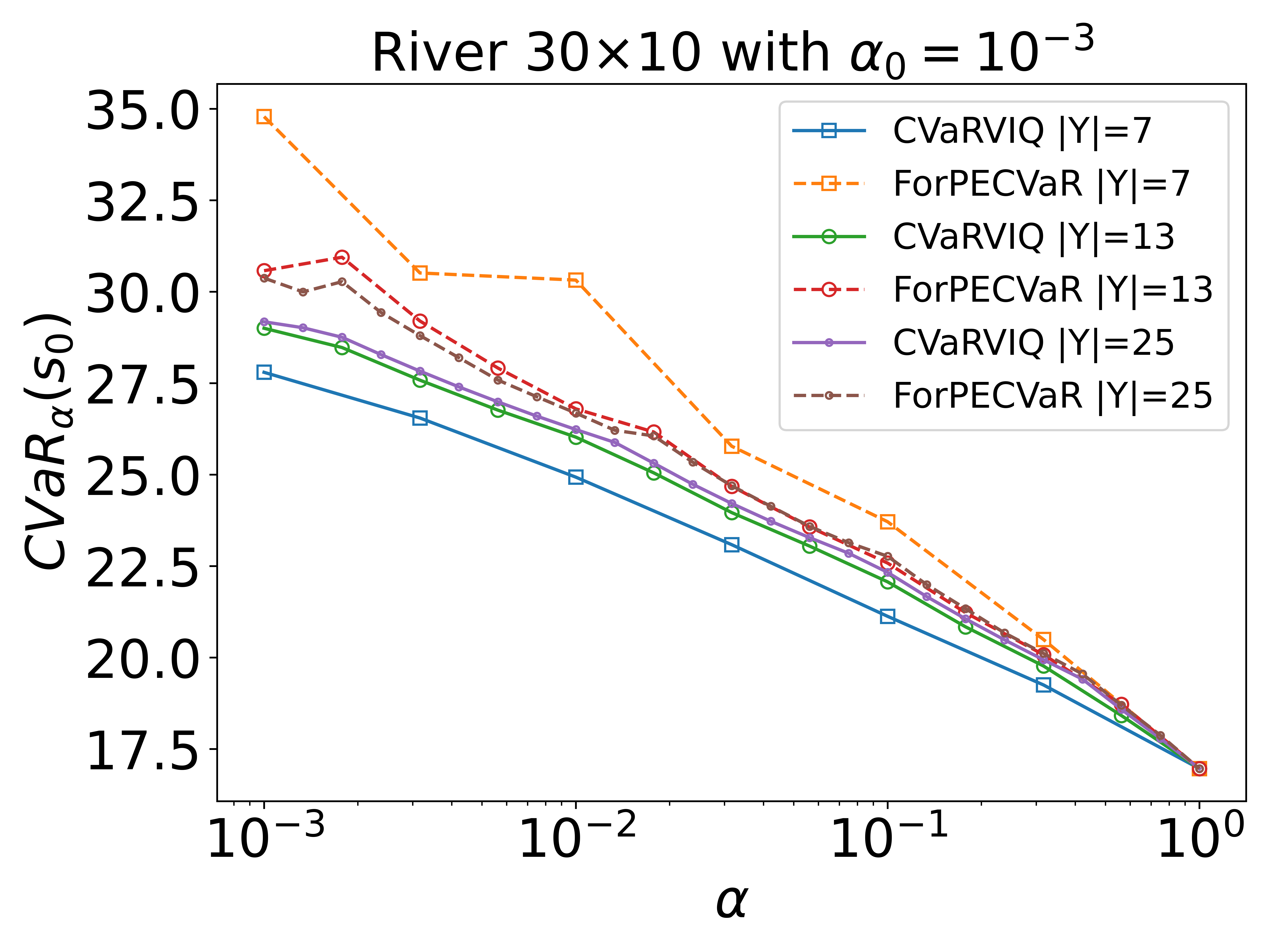}
\caption{\centering Approximate and exact values of \viq{} varying the number of atoms 
with a fixed $\alpha_0$.}
\label{fig:forpecvar_a0fixed}
\end{figure}

Fig. \ref{fig:forpecvar_a0fixed} shows the values for $\alpha_0 = 10^{-3}$ and $|Y|\in\{7,13,25\}$ for  Gridworld $14\times16$ and River $30\times10$.
The solid lines represent the approximate value found by CVaRVIQ and the dashed lines represent the policy evaluation of the \viq{} policy using \forpec{}. 
In Fig. \ref{fig:forpecvar_a0fixed} we can observe that with more atoms there is an increase in the approximate values and a decrease in the exact values so that the distance between approximate and exact values decreases, which is consistent with Theorem 7 of \cite{chow:risk-sensitive}. 
Considering all the experiments performed, the same behavior is observed.
However, in general, points closer to $\alpha_0$ have a greater distance between approximate and exact values than points closer to $1$.

\begin{figure}[!t]
\centering
  \includegraphics[width=.23\textwidth]{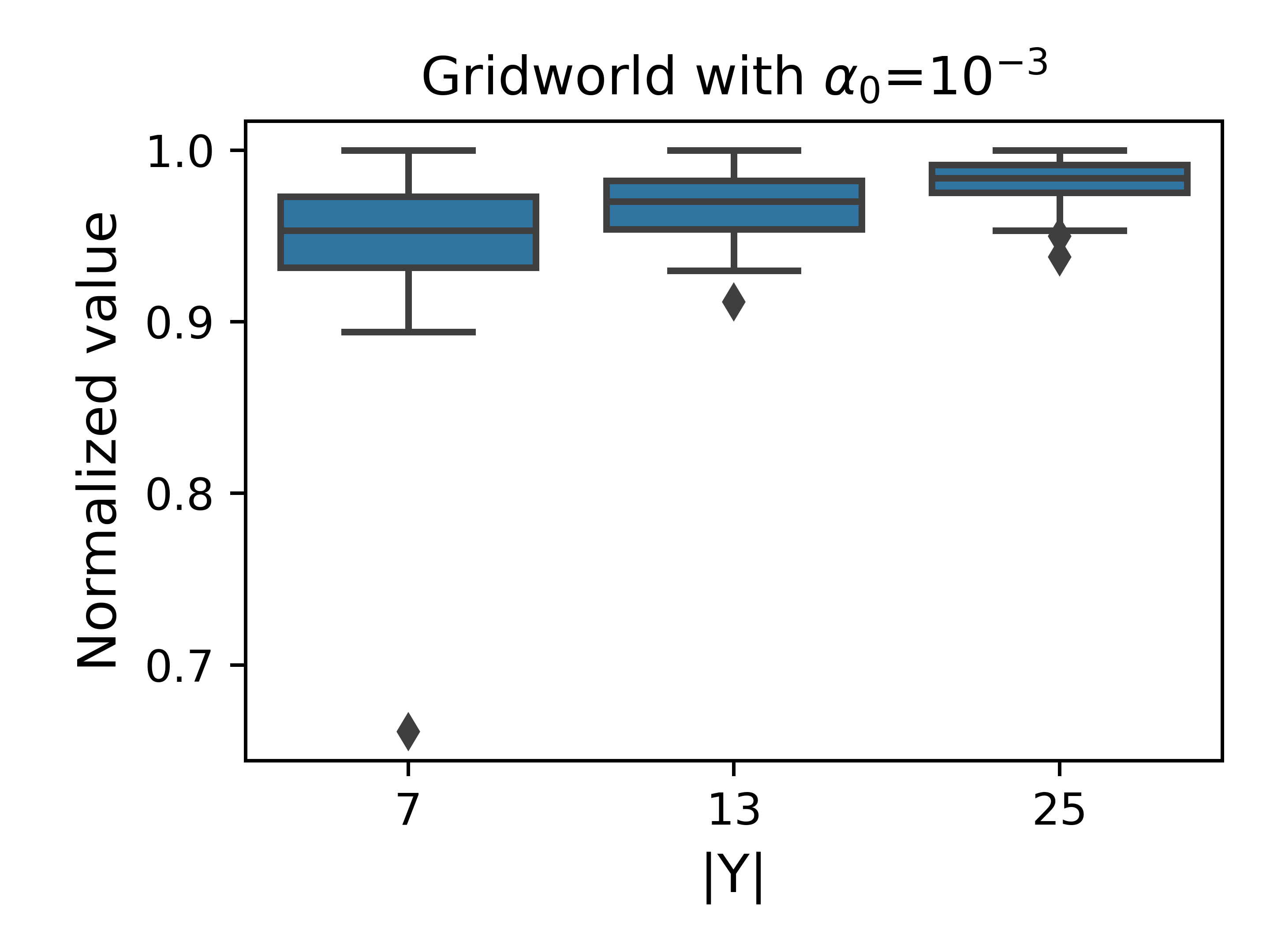}
  \includegraphics[width=.23\textwidth]{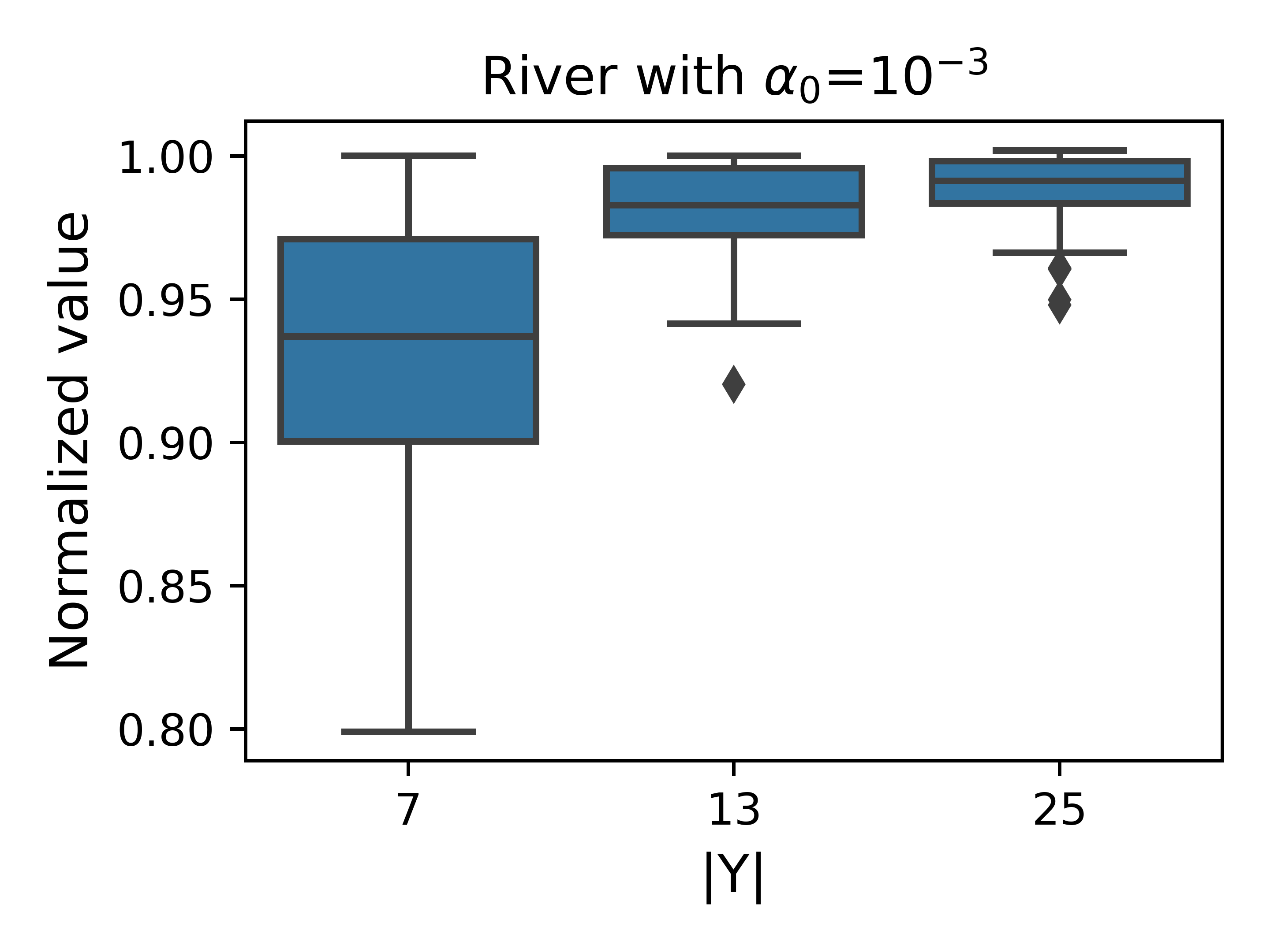}
\caption{\centering Boxplots with approximate values normalized by the exact values considering all problems of the Gridworld and River domains.}
\label{fig:boxplot_a0fixed}
\end{figure}

Fig. \ref{fig:boxplot_a0fixed} shows the boxplots with the results of all experiments performed for $\alpha_0=10^{-3}$.
Each boxplot has the values of the points of all experiments in the same domain with the same $\alpha_0$ and $|Y|$.
The approximate values were normalized by the exact value in order to characterize the distance between them.
Thus, normalized values closer to $1$ correspond to a better approximation of the \viq{} algorithm.
Boxplots from experiments with fewer atoms have fewer points.
For example, boxplots for $|Y|=7$ have 21 points (7 atoms for each of the 3 problems in each domain).

The boxplots show that 
the normalized values are closer to $1$ with more atoms, i.e., approximations with few points produce worse policies, since the normalized value are farther from $1$. Thus, we can observe that the value and policy of \viq{} are moving toward the optimal value when using more atoms.
The outliers in the boxplots correspond to experiments with atoms values closer to $\alpha_0$ regardless of the number of atoms used in the approximation.
This happens because of a limitation of the approximate algorithms, which cannot approximate $\alpha_0$ well as observed in Section \ref{subsec:values_alpha0}.

\subsection{Values of \viq{} varying $\alpha_0$}
\label{subsec:values_alpha0}

Since the distance between the approximate and exact values at $\alpha_0$ is substantial regardless of the number of atoms used, in this section, we investigate the use of an $\alpha_0$ smaller than the $\alpha_{target}$. 

\begin{figure}[!t]
\centering
  \includegraphics[width=.23\textwidth]{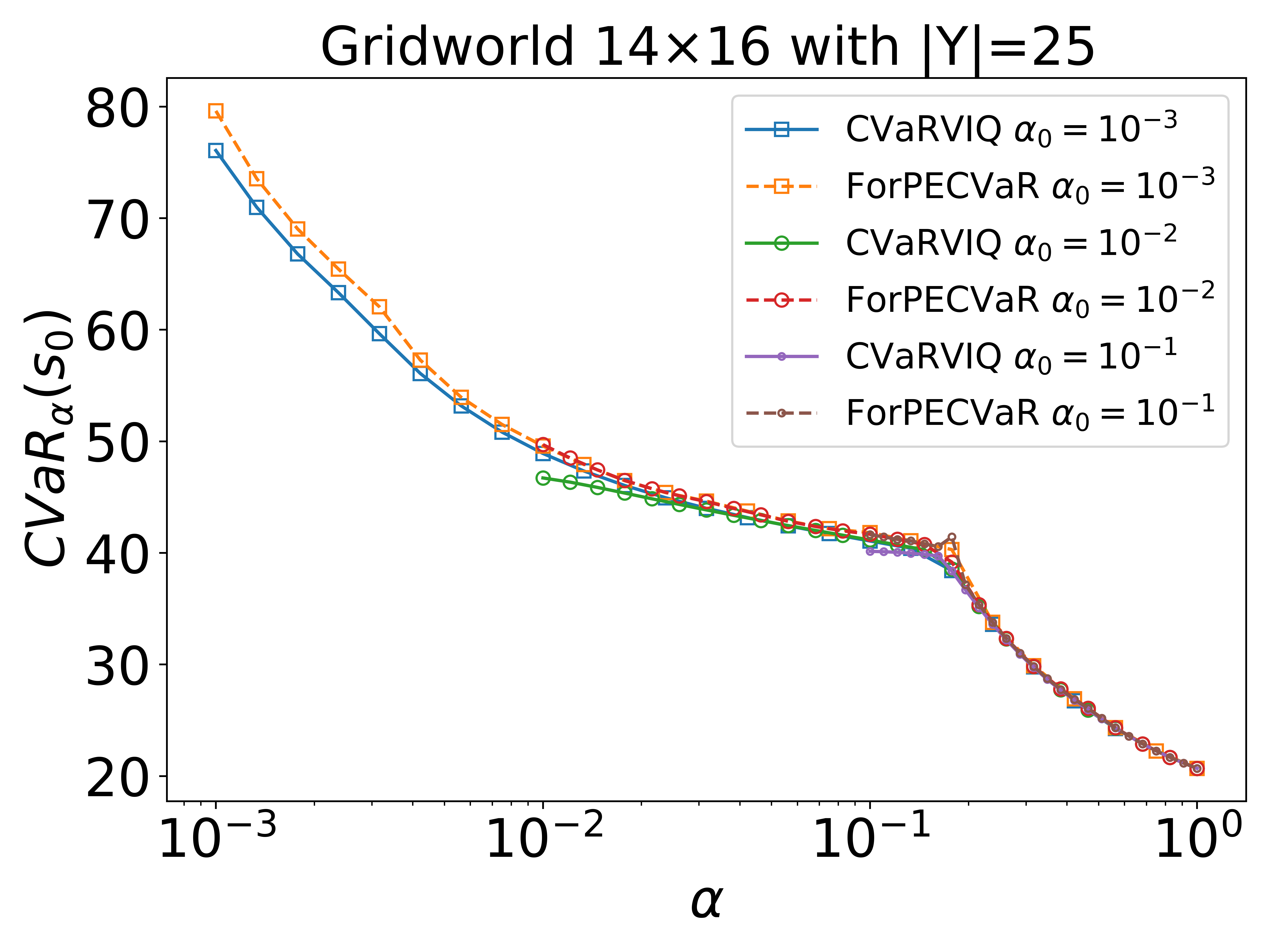}
  \includegraphics[width=.23\textwidth]{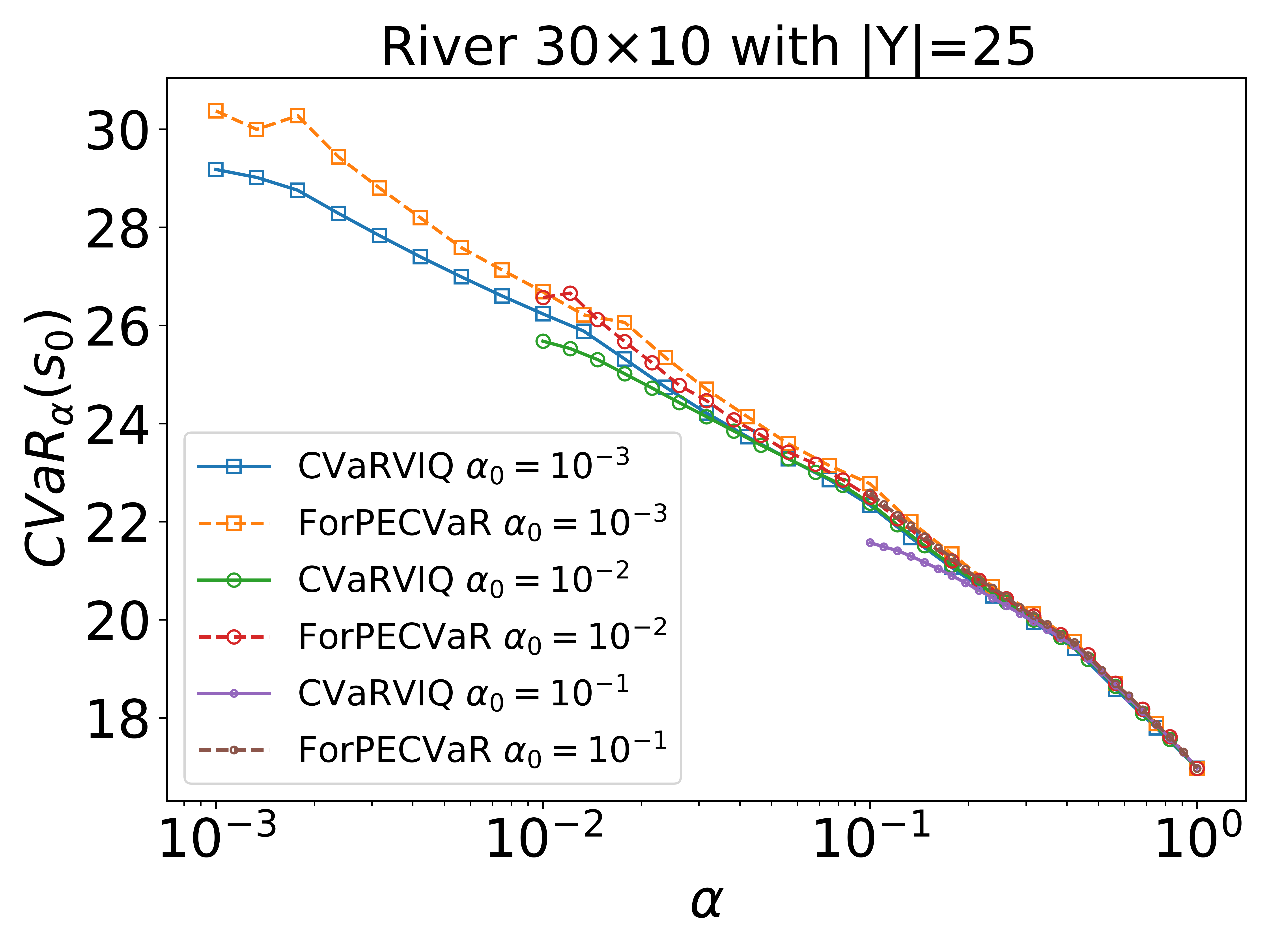}
\caption{\centering Approximate and exact values of \viq{} varying the $\alpha_0$ values
with a fixed $|Y|$.}
\label{fig:forpecvar_nYfixed}
\end{figure}

Fig. \ref{fig:forpecvar_nYfixed} shows the values of the experiments with $25$ atoms for the Gridworld $14\times16$ and River $30\times10$ problems, both varying $\alpha_0\in\{10^{-3}, 10^{-2}, 10^{-1}\}$.
In the Gridworld problem,
for $\alpha=10^{-2}$, the exact value for the experiment with $\alpha_0=10^{-3}$ (orange line) is smaller than the value for the experiment with $\alpha_0=10^{-2}$ (red line). Analogously, for $\alpha=10^{-1}$, the exact values  considering the experiments with $\alpha_0=10^{-3}$ and $\alpha_0=10^{-2}$ are better than with $\alpha_0=10^{-1}$.
In the River problem, the same behavior does not happen, i.e., the exact value for the experiment with $\alpha_0=10^{-3}$ (orange line) is not smaller than the value for the experiment with $\alpha_0=10^{-2}$ (red line). This happen because the number of approximation points was not enough for obtaining a good solution. However, considering more atoms, the behavior is  similar to the Gridworld results.

\begin{figure}[!t]
\centering
  \includegraphics[width=.23\textwidth]{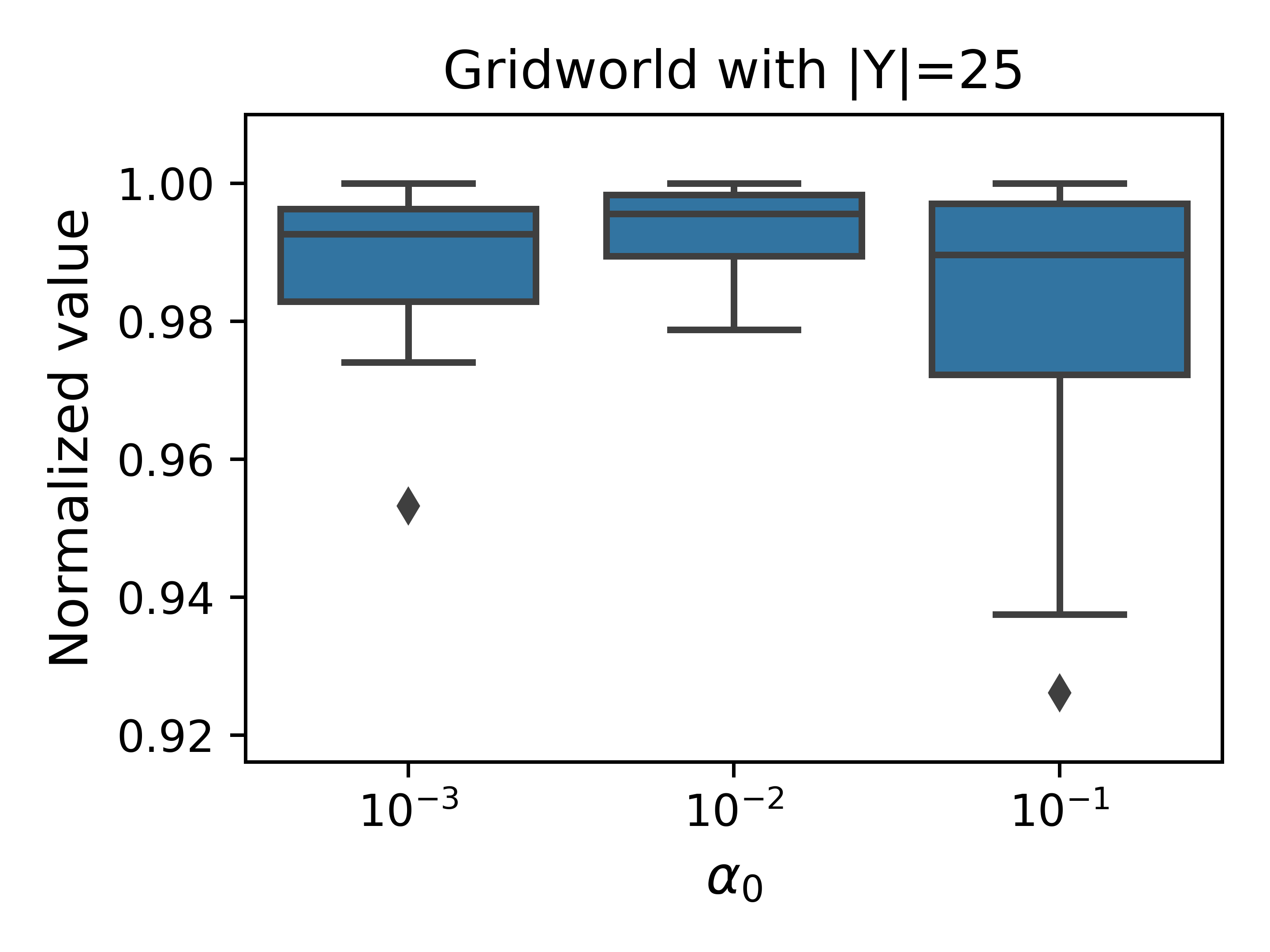}
  \includegraphics[width=.23\textwidth]{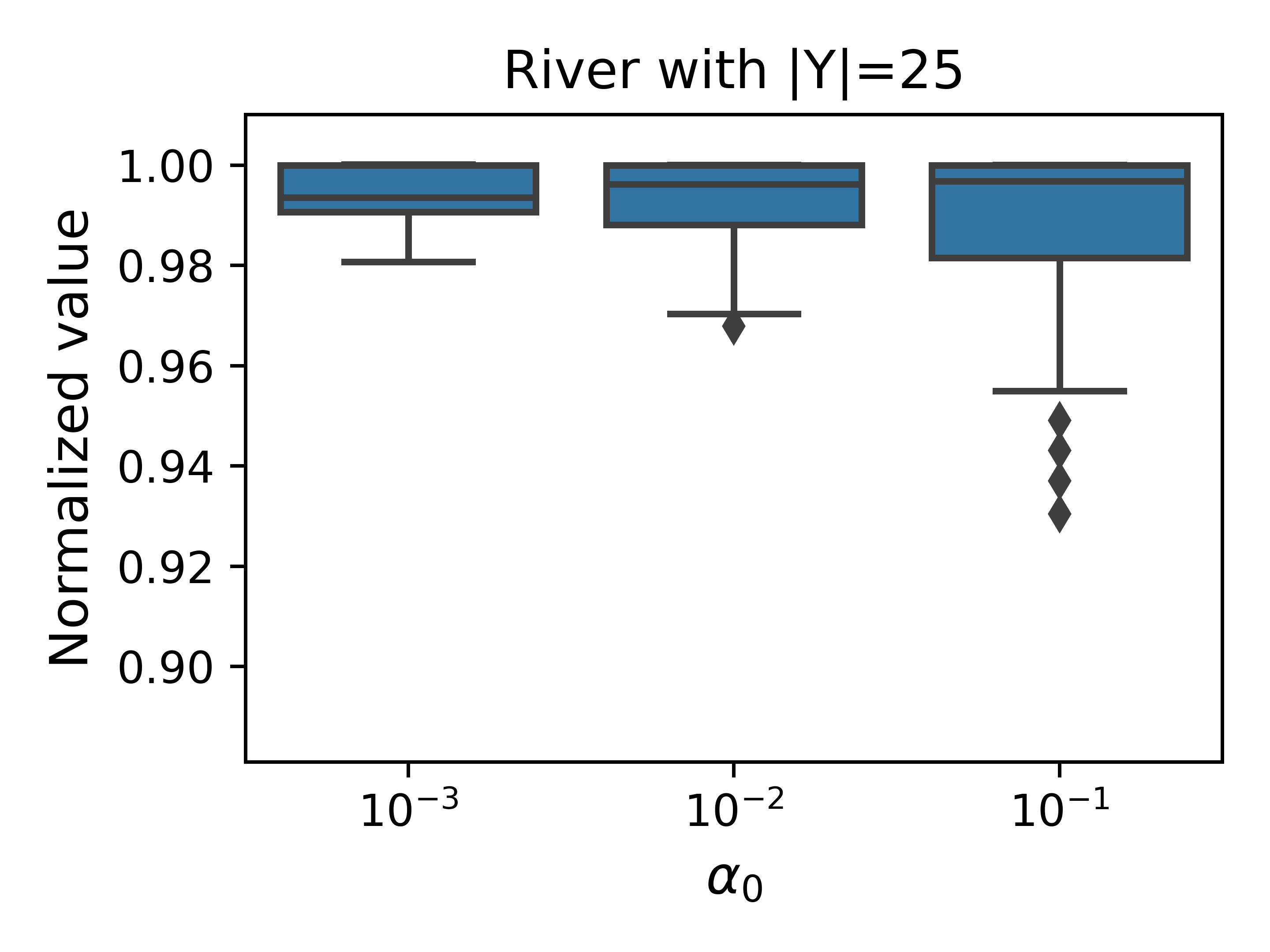}
\caption{\centering Boxplots with approximate values normalized by the exact values considering all problems of the Gridworld and River domains.} 
\label{fig:boxplot_nYfixed}
\end{figure}

The boxplots in Fig. \ref{fig:boxplot_nYfixed} show the approximate values normalized to the exact values for $\alpha_{target}=10^{-1}$. 
For each one of the $\alpha_0$, we selected the points of the respective $Y$ 
as $\widetilde{Y} = \{y | y \in Y \wedge y \geq \alpha_{target} \}$  where  $y_1\in Y$ is equal to $\alpha_0$.
Note that
for each $\alpha_0$ different points 
could be selected\footnote{For example, if $|Y|=7$ and $\alpha_{target}=10^{-1}$, then 
for $\alpha_0=10^{-2}$, the set of atoms $Y=\{$0.01, 0.022, 0.046, 0.1, 0.22, 0.46, 1$\}$, and the set $\widetilde{Y}=\{$0.1, 0.22, 0.46, 1$\}$.
For $\alpha_0=10^{-3}$, $Y=\{$0.001, 0.0032, 0.01, 0.032, 0.1, 0.32, 1$\}$ and $\widetilde{Y}=\{$0.1, 0.32, 1$\}$.}.
All points of each boxplot have the values of the experiments with the same $|Y|$ and $\alpha_0$ parameters. 
The experiments show that using a small number of atoms, $|Y|=7$ for example (boxplots not displayed because of the limit of space), the approximate values are far from the exact one. So first we can choose a suitable value of $|Y|$ (in these experiments it is better to choose 25). 
By setting $|Y|=25$, Fig. \ref{fig:boxplot_nYfixed} shows that to choose an appropriate value of $\alpha_0$ to approximate $\alpha_{target}$, it is necessary to choose a value not too close from $\alpha_{target}$. 

In the Gridworld and River domains, if we consider  $\alpha_0=\alpha_{target}=10^{-1}$ and $|Y|=25$, the minimum normalized value and the first quartile is worse than the other ones for other values of $\alpha_0$. 
In the Gridworld domain with $|Y|=25$, the first quartile for $\alpha_0= 10^{-2}$ is better than the first quartile for $\alpha_0= 10^{-3}$ and $\alpha_0= 10^{-1}$. 
In the River domain with $|Y|=25$, the first quartile for $\alpha_{0}=10^{-3}$ is better than the others.

Summing up, to obtain a good approximation for a problem with $\alpha_{target}$, first, we need to choose an adequate number of atoms and then choose $\alpha_0$ that is smaller than $\alpha_{target}$.

\subsection{Execution time of \forpec{}}

\begin{figure}[!t]
\centering
  \includegraphics[width=.23\textwidth]{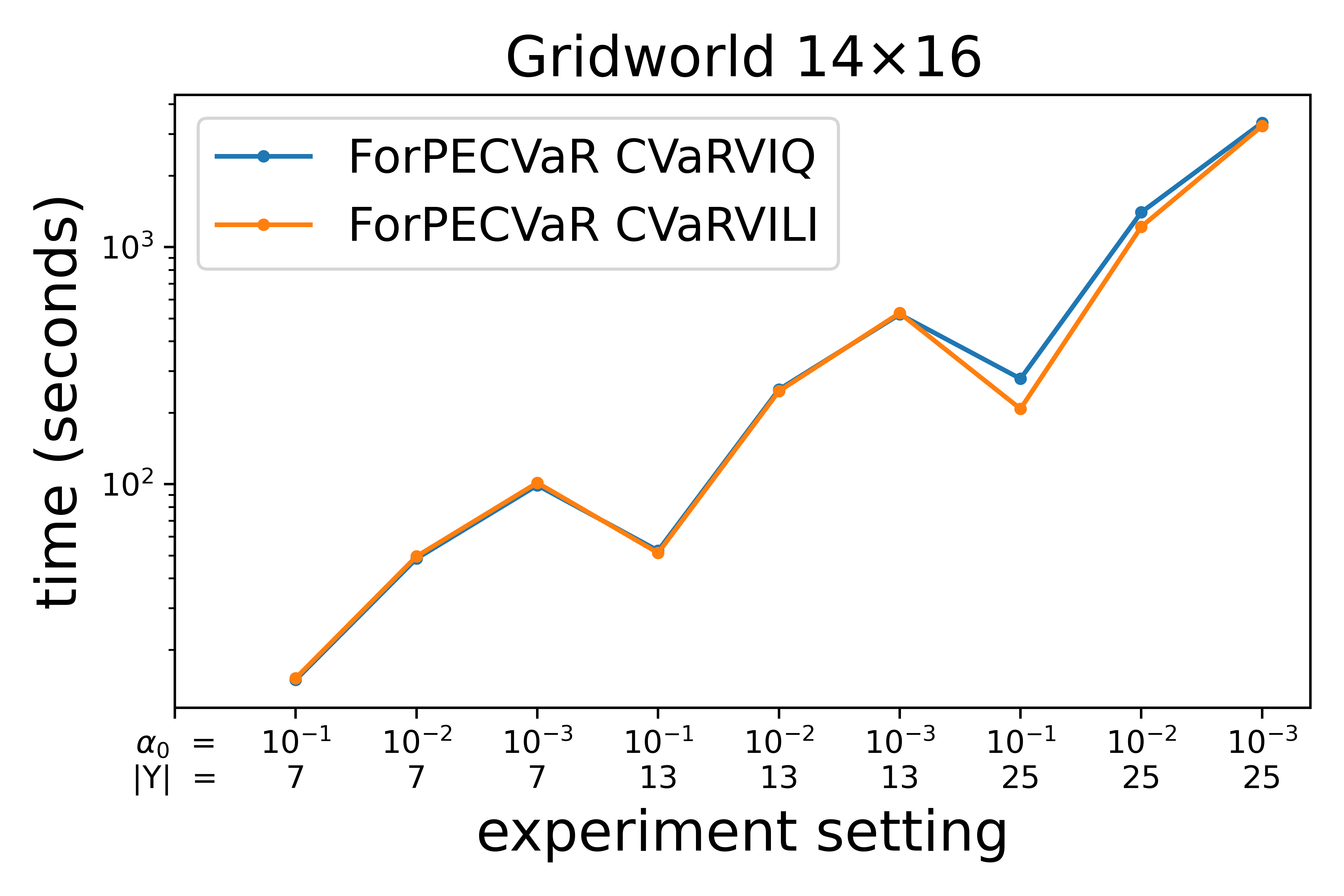}
  \includegraphics[width=.23\textwidth]{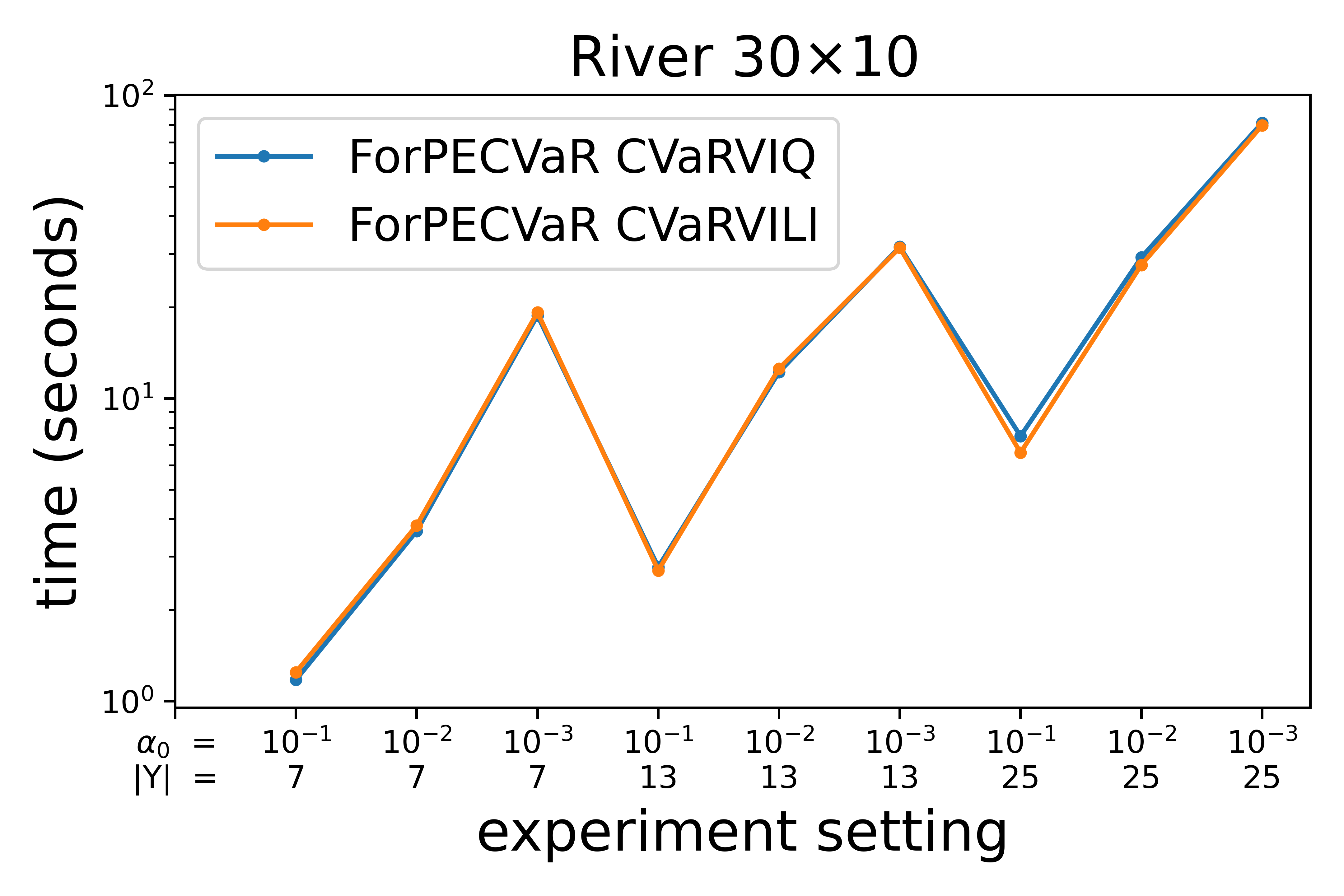}
\caption{\centering Execution time  of \forpec{} evaluation of \viq{} and \vili{} policies for the atom $\alpha_{target}=\alpha_0$.}
\label{fig:time_forpec_both}
\end{figure}

Fig. \ref{fig:time_forpec_both} shows the runtime of \forpec{} evaluation of \vili{} and \viq{} policies for Gridworld $14\times16$ and the River $30\times10$ problems in seconds for different settings varying $\alpha_{0}$ and $|Y|$ 
for the atom $\alpha_{target} = \alpha_0$.
For all experiments for a fixed $|Y|$, the lower the $\alpha_0$, the longer it takes to evaluate the policy.
This is because it is necessary to reach the goal state  with a higher probability.
When fixing $\alpha_0$, we see that with more atoms, the execution time is longer, because with more atoms, the approximation of the policy is better, and the policy will tend to take more safe actions, which will take more time to reach the goal state with the necessary probability.

The Monte Carlo simulation (MC) technique 
can be used to evaluate approximately policies.
We run 5 times MC to evaluate each of the 18  policies that correspond to the configurations of Fig. \ref{fig:time_forpec_both} with the same time spent by \forpec{} to evaluate the \viq{} policy. 
We calculate the difference between exact values computed by \forpec{} and approximated values computed by MC. The mean was 0.67 and the median was 0.27 for the Gridworld problem, and 0.45 and 0.23 for the River problem, respectively on average. The maximum difference was 11.82 for the Gridworld problem and 5.29 for the River problem on average.
Thus, MC can not get accurate evaluations of the performance of the policies 
considering the same time spent by ForPECVaR.

Table \ref{table:time_forpecvar} shows the
maximum execution time 
in seconds of \forpec{} evaluation of \viq{} and \vili{} policies for the atom $\alpha_{target}=\alpha_0$. 
In all cases, the execution time corresponds to the configuration of $|Y|=25$ and $\alpha_0=10^{-3}$, which are the highest number of atoms and the lowest confidence level tested.
Gridworld problems policies have more execution time because the path toward the goal is longer than in the River problems, and more trajectories are expanded in the evaluation.

\begin{table}[!h]
\centering
\begin{tabular}{|c|ccc|ccc|}
\hline
\begin{tabular}[c]{@{}c@{}}maximum\\ time\end{tabular} & \multicolumn{3}{c|}{Gridworld}                                                   & \multicolumn{3}{c|}{River}                                                         \\ \cline{2-7} 
(seconds)                                              & \multicolumn{1}{c|}{$5\times5$} & \multicolumn{1}{c|}{$8\times9$} & $14\times16$ & \multicolumn{1}{c|}{$10\times3$} & \multicolumn{1}{c|}{$16\times6$} & $30\times10$ \\ \hline
CVaRVILI                                               & \multicolumn{1}{c|}{6.79}       & \multicolumn{1}{c|}{122.52}      & 3244.18       & \multicolumn{1}{c|}{0.02}        & \multicolumn{1}{c|}{2.27}        & 79.67         \\ \hline
CVaRVIQ                                                & \multicolumn{1}{c|}{7.95}       & \multicolumn{1}{c|}{130.12}      & 3344.86      & \multicolumn{1}{c|}{0.02}       & \multicolumn{1}{c|}{2.29}        & 81.32         \\ \hline
\end{tabular}
\caption{\centering 
Maximum execution time of \forpec{} evaluation of \viq{} and \vili{} policies for %the atom 
$\alpha_{target}=\alpha_0$.}
\label{table:time_forpecvar}
\end{table}

\section{Related Work}
\label{sec:related}

CVaRVILI finds an optimal approximate policy for CVaR MDP problems using linear interpolation.
CVaRVIQ 
uses the connection between the $yCVaR_y$ and $VaR_y$ functions. 
The \forpec{} algorithm is able to evaluate the policy returned by \vili{} and \viq{} for problems with non-uniform costs.
Meggendorfer \cite{Meggendorfer22} also proposed an algorithm to evaluate this type of policy. However, this algorithm only works for problems with uniform cost. Additionally, the code  is not available and no experiments with this algorithm were performed in \cite{Meggendorfer22}.
ForPECVaR differs from this algorithm  because the  computed equations are different and it tracks the accumulated cost for each trajectory from the initial state independently. Each trajectory is added to a priority queue with respect to the accumulated cost and a heuristic can be used to improve the execution time of the algorithm.
Recently, a Value Iteration based algorithm was proposed to find the optimal value of  CVaR-SSPs \cite{Meggendorfer22}. 
Note that these algorithms proposed in \cite{Meggendorfer22} were independently developed from our proposal.

Rigter et al. \cite{rigter2022planning} formulate a lexicographic optimization problem that extends CVaRVILI and minimizes the expected cost subject to the constraint that the CVaR of the total cost is optimal.
They show that there are multiple policies that get the same optimal CVaR value. However, they need the VaR value in their lexicographic approach that is obtained through 
MC simulations of the optimal CVaR policy in their experiments. \forpec{} algorithm can be used by this algorithm since it also returns the exact VaR value.
In \cite{carpin:aversion} is defined a surrogate MDP problem to model a CVaR in the transient total cost MDP (similar to SSP). The solution to this problem approximates the optimal policy.

The CVaR criterion was also studied in the risk-sensitive reinforcement learning (RL) area
\cite{chow2014algorithms, chow2018risk,tamar2015optimizing, prashanth2014policy,stanko2019risk, keramati2020being,pmlr-v100-tang20a}.
Our proposal does not evaluate these policies as it needs state transitions to assess them.
Another way to consider the CVaR criterion is through the use of constrained MDPs problems that considers a user-defined CVaR threshold as a constraint of an expected value optimization problem \cite{chow2014algorithms,borkar2014risk,prashanth2014policy,chow2018risk}.
\cite{rigter2022planning}.
The policies found with the CVaR-constrained problems can also be evaluated by \forpec{} as long as the transitions of the model are known.

\section{Conclusion}
\label{sec:conclusion}

Given the existence of many algorithms with approximation to solve CVaR MDPs problems, it is important to have exact algorithms to evaluate them and the influence of their parameters.
In this paper, we have presented \forpec{}, an exact algorithm to evaluate any CVaR policy 
with a forward approach. 
In addition to the CVaR value, \forpec{} also calculates the exact VaR value of the policies
that can be used by other algorithms such as the algorithm proposed in \cite{rigter2022planning}.
Our experimental evaluation has demonstrated that the approximate algorithms \vili{} and \viq{} return similar policies and values, but the second has a better execution time.
The exact evaluation of the \viq{} policy shows a limitation of the algorithms analyzed in relation to the approximation of the values and policies closest to the minimum confidence level $\alpha$.
We also showed that the simple approach of MC can not get accurate evaluations of policies considering the same time used by ForPECVaR.

\section*{Acknowledgments}
This study was financed in part by the Coordenação de Aperfeiçoamento de Pessoal de Nível Superior – Brasil (CAPES) – Finance Code 001 
and the Center for Artificial Intelligence
(C4AI-USP), with support by 
FAPESP (grant \#2019/07665-4) and by the IBM Corporation.

%%%%%%%%%%%%%%%%%%%%%%%%%%%%%%%%%%%%%%%%%%%%%%%%%%%%%%%%%%%%%%%%%%%%%%%%

%%% The next two lines define, first, the bibliography style to be 
%%% applied, and, second, the bibliography file to be used.
%\newpage
\bibliographystyle{ACM-Reference-Format} 
\bibliography{main}

%%%%%%%%%%%%%%%%%%%%%%%%%%%%%%%%%%%%%%%%%%%%%%%%%%%%%%%%%%%%%%%%%%%%%%%%

\end{document}